\documentclass[10pt,twocolumn,letterpaper]{article}

\usepackage{cvpr}
\usepackage{times}
\usepackage{epsfig}
\usepackage{graphicx}
\usepackage{amsmath}
\usepackage{amssymb}
\usepackage{rotating}
\usepackage{color}
\usepackage{colortbl}
\usepackage{framed}
\usepackage{subcaption}
\usepackage{arydshln}
\usepackage{multirow}
\usepackage{booktabs}
\usepackage{bm}
\usepackage{enumitem}

\definecolor{Gray}{gray}{0.9}

\usepackage[pagebackref=true,breaklinks=true,letterpaper=true,colorlinks,bookmarks=false]{hyperref}

\cvprfinalcopy

\def \ie {\emph{i.e.}}

\def \etal {\emph{et al.}}

\def \ours {Art2Real\xspace}

\newcommand{\tit}[1]{\smallskip\noindent\textbf{#1.}}
\newcommand{\tinytit}[1]{\noindent\textbf{#1.}}

\begin{document}

\title{\textit{\ours}: Unfolding the Reality of Artworks\\ via Semantically-Aware Image-to-Image Translation}

\author{Matteo Tomei, Marcella Cornia, Lorenzo Baraldi, Rita Cucchiara \\
University of Modena and Reggio Emilia\\
{\tt\small \{name.surname\}@unimore.it}
}

\maketitle

\begin{abstract}
The applicability of computer vision to real paintings and artworks has been rarely investigated, even though a vast heritage would greatly benefit from techniques which can understand and process data from the artistic domain. This is partially due to the small amount of annotated artistic data, which is not even comparable to that of natural images captured by cameras.
In this paper, we propose a semantic-aware architecture which can translate artworks to photo-realistic visualizations, thus reducing the gap between visual features of artistic and realistic data. Our architecture can generate natural images by retrieving and learning details from real photos through a  similarity matching strategy which leverages a weakly-supervised semantic understanding of the scene. Experimental results show that the proposed technique leads to increased realism and to a reduction in domain shift, which improves the performance of pre-trained architectures for classification, detection, and segmentation. Code is publicly available at: \url{https://github.com/aimagelab/art2real}.
\end{abstract}


\section{Introduction}
\label{sec:introduction}

Our society has inherited a huge legacy of cultural artifacts from past generations: buildings, monuments, books, and exceptional works of art. While this heritage would benefit from algorithms which can automatically understand its content, computer vision techniques have been rarely adapted to work in this domain.

One of the reasons is that applying state of the art techniques to artworks is rather difficult, and often brings poor performance. This can be motivated by the fact that the visual appearance of artworks is different from that of photo-realistic images, due to the presence of brush strokes, the creativity of the artist and the specific artistic style at hand. As current vision pipelines exploit large datasets consisting of natural images, learned models are largely biased towards them. The result is a gap between high-level convolutional features of the two domains, which leads to a decrease in performance in the target tasks, such as classification, detection or segmentation.

\begin{figure}[t]
    \centering
    \includegraphics[width=0.98\linewidth]{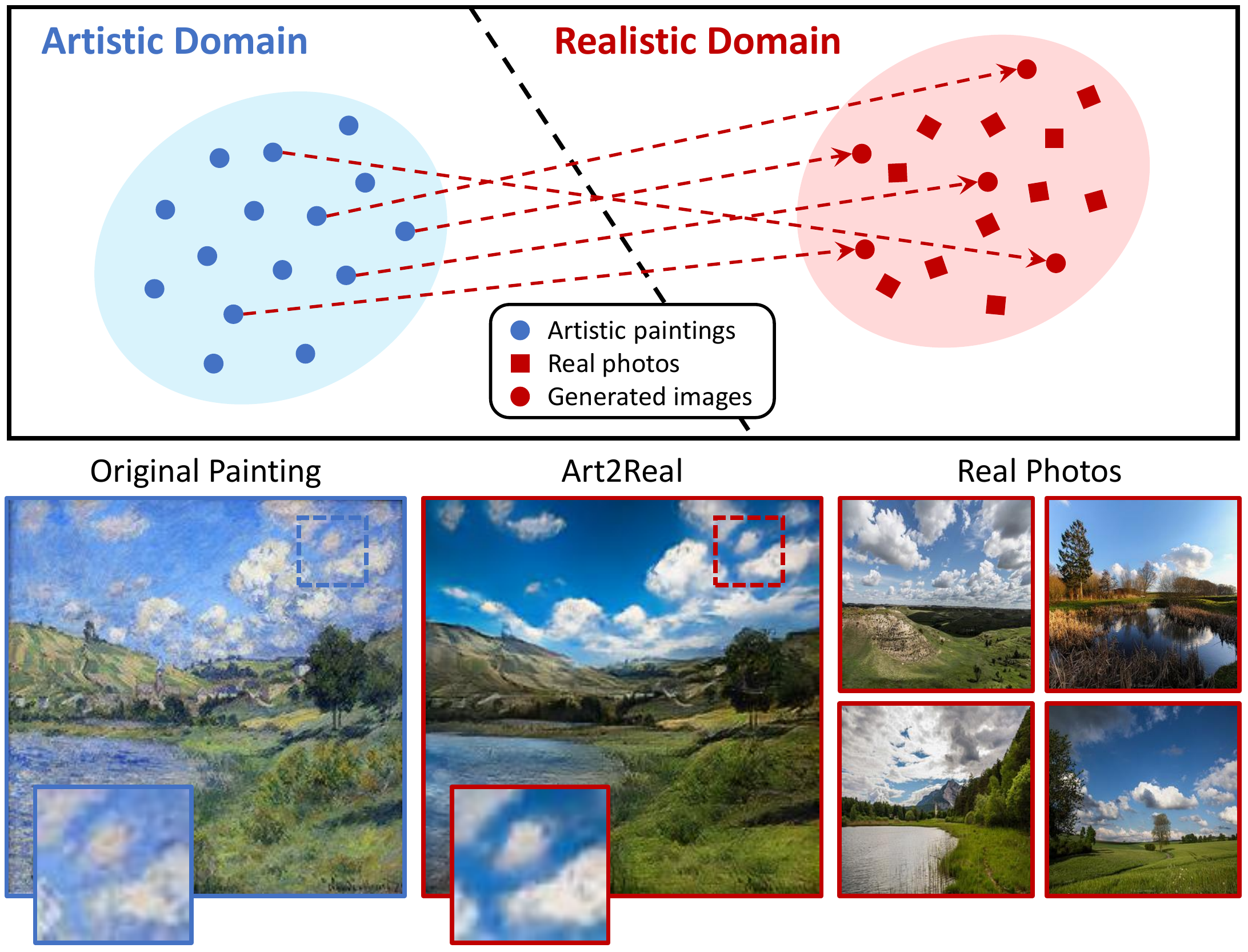}
    \caption{We present \textit{\ours}, an architecture which can reduce the gap between the distributions of visual features from artistic and realistic images, by translating paintings to photo-realistic images.}
    \label{fig:art_real}
    \vspace{-.3cm}
\end{figure}

This paper proposes a solution to the aforementioned problem that avoids the need for re-training neural architectures on large-scale datasets containing artistic images. In particular, we propose an architecture which can reduce the shift between the feature distributions from the two domains, by translating artworks to photo-realistic images which preserve the original content. A sample of this setting is depicted in Fig. \ref{fig:art_real}.

As paired training data is not available for this task, we revert to an unpaired image-to-image translation setting~\cite{zhu2017unpaired}, in which images can be translated between different domains while preserving some underlying characteristics. In our \textit{art-to-real} scenario, the first domain is that of paintings while the second one is that of natural images. The shared characteristic is that they are two different visualizations of the same class of objects, for example, they both represent landscapes.

In the translation architecture that we propose, new photo-realistic images are obtained by retrieving and learning from existing details of natural images and exploiting a weakly-supervised semantic understanding of the artwork. To this aim, a number of memory banks of realistic patches is built from the set of photos, each containing patches from a single semantic class in a memory-efficient representation. By comparing generated and real images at the patch level, in a multi-scale manner, we can then drive the training of a generator network which learns to generate photo-realistic details, while preserving the semantics of the original painting. 
As performing a semantic understanding of the original painting would create a chicken-egg problem, in which unreliable data is used to drive the training and the generation, we propose a strategy to update the semantic masks during the training, leveraging the partial convergence of a cycle-consistent framework.

We apply our model to a wide range of artworks which include paintings from different artists and styles, landscapes and portraits. Through experimental evaluation, we show that our architecture can improve the realism of translated images when compared to state of the art unpaired translation techniques. This is evaluated both qualitatively and quantitatively, by setting up a user study. Furthermore, we demonstrate that the proposed architecture can reduce the domain shift when applying pre-trained state of the art models on the generated images.

\tit{Contributions}
To sum up, our contributions are as follows:
\begin{itemize}[noitemsep,topsep=0pt]
    \item We address the domain gap between real images and artworks, which prevents the understanding of data from the artistic domain. To this aim, we propose a network which can translate paintings to photo-realistic generated images.
    \item The proposed architecture is based on the construction of efficient memory banks, from which realistic details can be recovered at the patch level. Retrieved patches are employed to drive the training of a cycle-consistent framework and to increase the realism of generated images. This is done in a semantically aware manner, exploiting segmentation masks computed on artworks and generated images during the training.
    \item We show, through experimental results in different settings, improved realism with respect to state of the art approaches for image translation, and an increase in the performance of pre-trained models on generated data.
\end{itemize}

\section{Related work}
\label{sec:related}

\tinytit{Image-to-image translation} Generative adversarial networks have been applied to several conditional image generation problems, ranging from image inpainting~\cite{pathak2016context,yeh2017semantic,yu2018generative,yan2018shift} and super-resolution~\cite{ledig2017photo} to video prediction~\cite{mathieu2016deep,villegas2017decomposing,walker2017pose,liang2017dual} and text to image synthesis~\cite{reed2016generative,reed2016learning,zhang2017stackgan,xu2018attngan}.
Recently, a line of work on image-to-image translation has emerged, in both paired~\cite{isola2017image,sangkloy2017scribbler} and unpaired settings~\cite{zhu2017unpaired,kim2017learning,liu2017unsupervised,tomei2018monet}. Our task belongs to the second category, as the translation of artistic paintings to photo-realistic images cannot be solved by exploiting supervised methods.   

Zhu~\etal~\cite{zhu2017unpaired} proposed the Cycle-GAN framework, which learns a translation between domains by exploiting a cycle-consistent constraint that guarantees the consistency of generated images with respect to original ones. On a similar line, Kim~\etal~\cite{kim2017learning} introduced a method for preserving the key attributes between the input and the translated image, while preserving a cycle-consistency criterion. On the contrary, Liu~\etal~\cite{liu2017unsupervised} used a combination of generative adversarial networks, based on CoGAN~\cite{liu2016coupled}, and variational auto-encoders. While all these methods have achieved successful results on a wide range of translation tasks, none of them has been specifically designed, nor applied, to recover photo-realism from artworks.

A different line of work is multi-domain image-to-image translation~\cite{choi2018stargan,anoosheh2018combogan,yang2018crossing}: here, the same model can be used for translating images according to multiple attributes (\ie, hair color, gender or age).
Other methods, instead, focus on diverse image-to-image translation, in which an image can be translated in multiple ways by encoding different style properties of the target distribution~\cite{zhu2017toward,huang2018munit,DRIT}. However, since these methods typically depend on domain-specific properties, they are not suitable for our setting as realism is more important than diversity.

\begin{figure*}[t]
    \centering
    \includegraphics[width=0.98\linewidth]{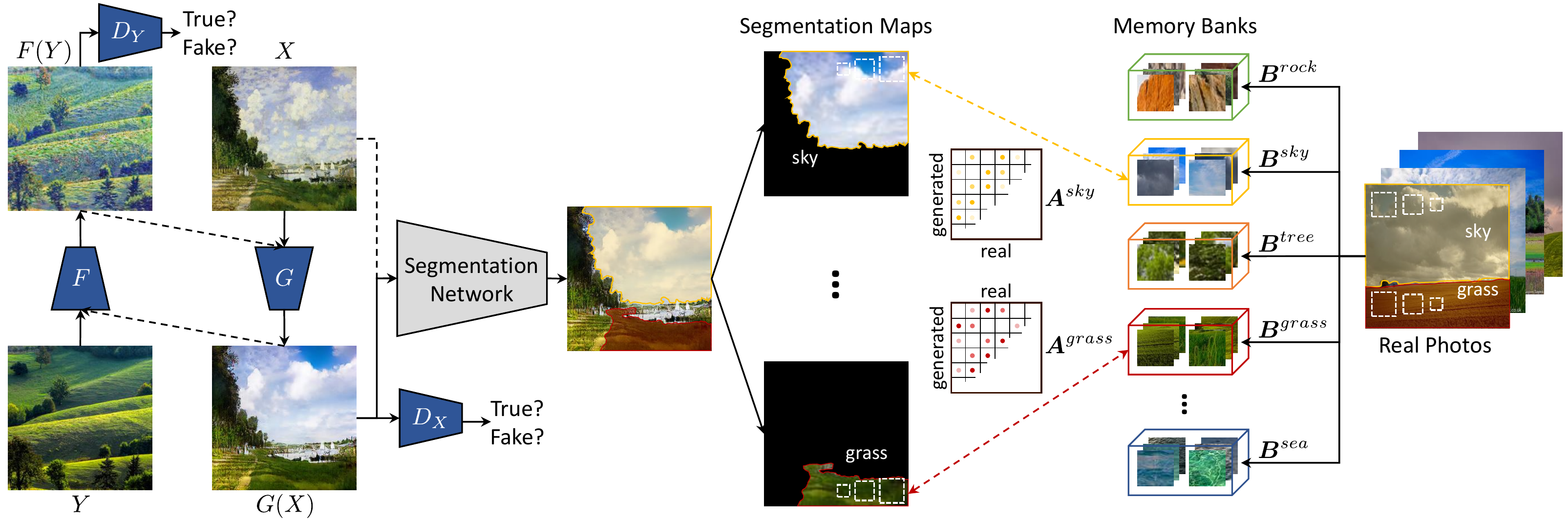}
    \caption{Overview of our \textit{\ours} approach. A painting is translated to a photo-realistic visualization by forcing a matching with patches from real photos. This is done in a semantically-aware manner, by building class-specific memory banks of real patches $\bm{B}^c$, and pairing generated and real patches through affinity matrices $\bm{A}^c$, according to their semantic classes. Segmentation maps are computed either from the original painting or the generated image as the training proceeds.}
    \label{fig:model}
    \vspace{-.3cm}
\end{figure*}

\tit{Neural style transfer}
Another way of performing image-to-image translation is that of neural style transfer methods~\cite{gatys2015neural,gatys2016image,johnson2016perceptual,huang2017arbitrary,sanakoyeu2018style}, in which a novel image is synthesized by combining the content of one image with the style of another, typically a painting. In this context, the seminal work by Gatys~\etal~\cite{gatys2015neural,gatys2016image} proposed to jointly minimize a content loss to preserve the original content, and a style reconstruction loss to transfer the style of a target artistic image. The style component is encoded by exploiting the Gram matrix of activations coming from a pre-trained CNN. Subsequent methods have been proposed to address and improve different aspects of style transfer, including the reduction of the computational overhead~\cite{johnson2016perceptual,li2016precomputed,ulyanov2016texture}, the improvement of the generation quality~\cite{gatys2017controlling,chen2016fast,wang2017multimodal,jing2018stroke,sanakoyeu2018style} and diversity~\cite{li2017diversified,ulyanov2017improved}. Other works have concentrated on the combination of different styles~\cite{chen2017stylebank}, and the generalization to previously unseen styles~\cite{li2017universal,ghiasi2017exploring,shen2018neural}. All these methods, while being effective on transferring artistic styles, show poor performance in the opposite direction.

\section{Proposed approach}
\label{sec:approach}

Our goal is to obtain a photo-realistic representation of a painting. The proposed approach explicitly guarantees the realism of the generation and a semantic binding between the original artwork and the generated picture.
To increase the realism, we build a network which can copy from the details of real images at the patch level. Further, to reinforce the semantic consistency before and after the translation, we make use of a semantic similarity constraint: each patch of the generated image is paired up with similar patches of the same semantic class extracted from a memory bank of realistic images. The training of the network aims at maximizing this similarity score, in order to reproduce realistic details and preserve the original scene. An overview of our model is presented in Fig.~\ref{fig:model}.

\subsection{Patch memory banks}
\label{sec:patch_memory_banks}
Given a semantic segmentation model, we define a pre-processing step with the aim of building the memory banks of patches which will drive the generation. Each memory bank $\bm{B}^c$ is tied to a specific semantic class $c$, in that it can contain only patches which belong to its semantic class.
To define the set of classes, and  semantically understand the content of an image, we adopt the weakly-supervised segmentation model from Hu~\etal~\cite{hu2018learning}: in this approach, a network is trained to predict semantic masks from a large set of categories, by leveraging the partial supervision given by detections. We also define an additional background memory bank, to store all patches which do not belong to any semantic class.


Following a sliding-window policy, we extract fixed-size RGB patches from the set of real images and put them in a specific memory $\bm{B}^c$, according to the class label $c$ of the mask in which they are located. Since a patch might contain pixels which belong to a second class label or the background class, we store in $\bm{B}^c$ only patches containing at least $20\%$ pixels from class $c$.

\begin{figure}[t]
    \centering
    \includegraphics[width=\linewidth]{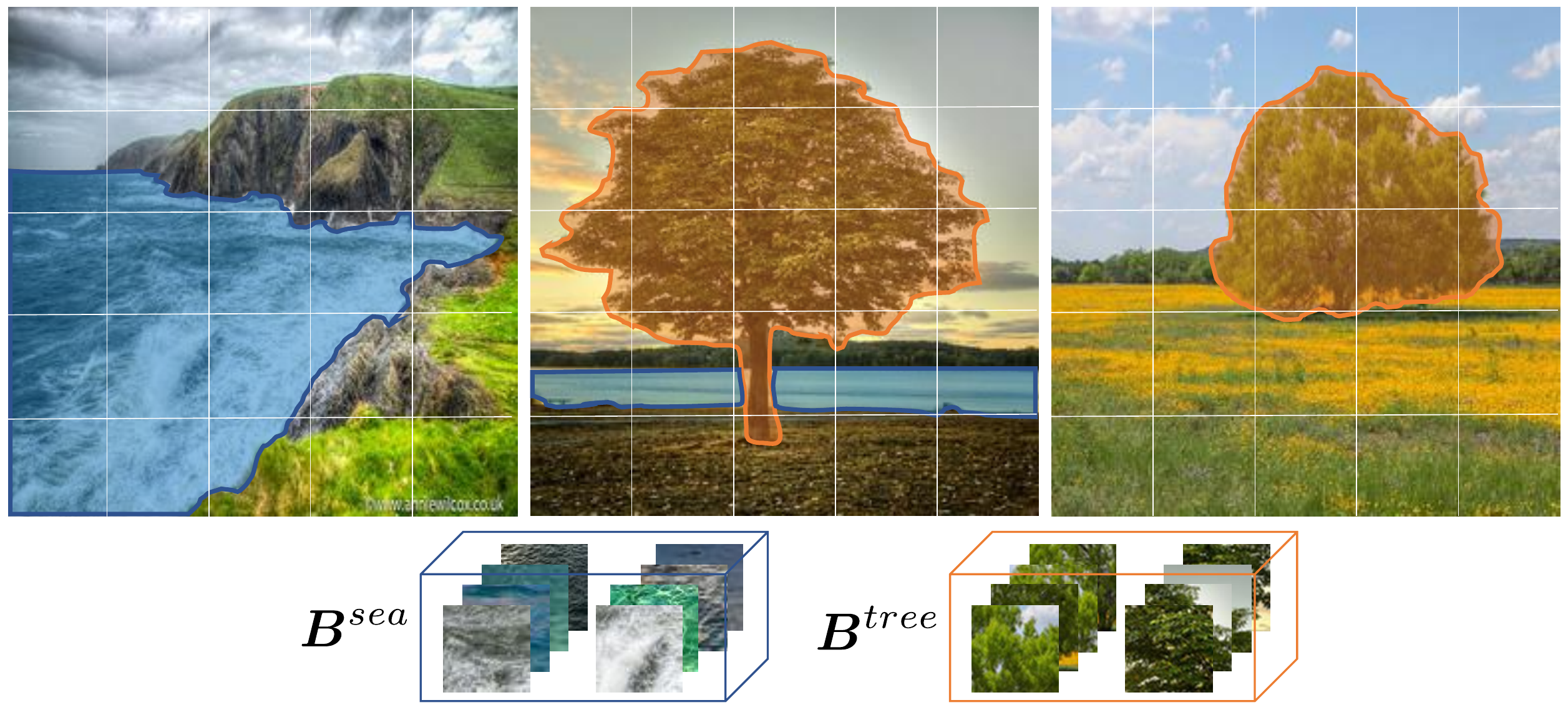}
    \caption{Memory banks building. A segmentation model~\cite{hu2018learning} computes segmentation masks for each realistic image in the dataset, then RGB patches belonging to the same semantic class are placed in the same memory bank.}
    \label{fig:seg_tree}
    \vspace{-.3cm}
\end{figure}

Therefore, we obtain a number of memory banks equal to the number of different semantic classes found in the dataset, plus the background class, where patches belonging to the same class are placed together (Fig.~\ref{fig:seg_tree}). Also, semantic information from generated images is needed: since images generated at the beginning of the training are less informative, we first extract segmentation masks from the original paintings. As soon as the model starts to generate meaningful images, we employ the segmentation masks obtained on generated images. 


\subsection{Semantically-aware generation}
The unpaired image-to-image translation model that we propose maps images belonging to a domain $X$ (that of artworks) to images belonging to a different domain $Y$ (that of natural images), preserving the overall content. Suppose we have a generated realistic image $G(x)$ at each training step, produced by a mapping function $G$ which starts from an input painting $x$. We adopt the previously obtained memory banks of realistic patches and the segmentation masks of the paintings in order to both enhance the realism of the generated details and keep the semantic content of the painting.

\tit{Pairing similar patches in a meaningful way}
At each training step, $G(x)$ is split in patches as well, maintaining the same stride and patch size used for the memory banks. Reminding that we have the masks for all the paintings, we denote a mask of the painting $x$ with class label $c$ as $\bm{M}^c_x$. We retrieve all masks $\bm{M}_x$ of the painting $x$ from which $G(x)$ originates, and assign each generated patch to the class label $c$ of the mask $\bm{M}^c_x$ in which it falls. If a patch belongs to different masks, it is also assigned to multiple classes. Then, generated patches assigned to a specific class $c$ are paired with similar realistic patches in the memory bank $\bm{B}^c$, \ie~the bank containing realistic patches with class label $c$. Given realistic patches belonging to $\bm{B}^c$, $\bm{B}^c=\{b^c_j\}$ and the set of generated patches with class label $c$, $\bm{K}^c=\{k^c_i\}$, we center both sets with respect to the mean of patches in $\bm{B}^c$, and we compute pairwise cosine distances as follows:
\begin{equation}
\label{eq:dist}
d_{ij}^c = \left(1-\frac{(k^c_i-\mu^c_b)\cdot(b^c_j-\mu^c_b)}{\left\lVert k^c_i-\mu^c_b \right\rVert_2 \left\lVert b^c_j-\mu^c_b \right\rVert_2}\right)
\end{equation} 
where $\mu^c_b = \frac{1}{N_c}\sum_jb^c_j$, being $N_c$ the number of patches in memory bank $\bm{B}^c$. We compute a number of distance matrices equal to the number of semantic classes found in the original painting $x$. Pairwise distances are subsequently normalized as follows:
\begin{equation}
\label{eq:normdist}
\tilde{d}_{ij}^c = \frac{d_{ij}^c}{\min_ld_{il}^c + \epsilon} \text{, where } \epsilon=1e-5
\end{equation}
and pairwise affinity matrices are computed by applying a row-wise softmax normalization:
\begin{equation}
\label{eq:aff}
\bm{A}_{ij}^c = \frac{\exp(1-\tilde{d}_{ij}^c/h)}{\sum_l \exp(1-\tilde{d}_{il}^c/h)} = 
\begin{cases}
    \approx 1 & \text{if } \tilde{d}_{ij}^c \ll \tilde{d}_{il}^c \text{ } \forall \text{ } l \neq j \\
    \approx 0 & \text{otherwise}
\end{cases}
\end{equation}
where $h>0$ is a bandwidth parameter. Thanks to the softmax normalization, each generated patch $k_i^c$ will have a high-affinity degree with the nearest real patch and with other not negligible near patches. Moreover, affinities are computed only between generated and artistic patches belonging to the same semantic class.

\tit{Approximate affinity matrix}
Computing the entire affinity matrix would require an intractable computational overhead, especially for classes with a memory bank containing millions of patches. In fact matrix $\bm{A}^c$ has as many rows as the number of patches of class $c$ extracted from $G(x)$ and as many columns as the number of patches contained in the memory bank $\bm{B}^c$.

To speed up the computation, we build a suboptimal Nearest Neighbors index $\bm{I}^c$ for each memory bank. When the affinity matrix for a class $c$ has to be computed, we conduct a $k$-NN search through $\bm{I}^c$ to get the $k$ nearest samples of each generated patch $k_i^c$. In this way, $\bm{A}^c$ will be a sparse matrix with at most as many columns as $k$ times the number of generated patches of class $c$. The Softmax in Eq.~\ref{eq:aff} ensures that the approximated version of the affinity matrix is very close to the exact one if the $k$-NN searches through the indices are reliable. We adopt inverted indexes with exact post-verification, implemented in the Faiss library~\cite{JDH17}. Patches are stored with their RGB values when memory banks have less than one million vectors; otherwise, we use a PCA pre-processing step to reduce their dimensionality, and scalar quantization to limit the memory requirements of the index.

\tit{Maximizing the similarity}
A contextual loss~\cite{mechrez2018learning} for each semantic class in $\bm{M}_x$ aims to maximize the similarity between couples of patches with high affinity value:
\begin{equation}
\label{eq:cx_class}
\mathcal{L}_{CX}^c(\bm{K}^c, \bm{B}^c) = -\log\left(\frac{1}{N_K^c}\left(\sum_i \max_j \bm{A}_{ij}^c\right)\right)
\end{equation}
where $N_K^c$ is the cardinality of the set of generated patches with class label $c$. Our objective is the sum of the previously computed single-class contextual losses over the different classes found in $\bm{M}_x$:
\begin{equation}
\label{eq:cx}
\mathcal{L}_{CX}(\bm{K}, \bm{B}) = \sum_{c}-\log\left(\frac{1}{N_K^c}\left(\sum_i \max_j \bm{A}_{ij}^c\right)\right)
\end{equation}
where $c$ assumes all the class label values of masks in $\bm{M}_x$. Note that masks in $\bm{M}_x$ are not constant during training: at the beginning, they are computed on paintings, then they are regularly extracted from $G(x)$.

\tit{Multi-scale variant}
To enhance the realism of generated images, we adopt a multi-scale variant of the approach, which considers different sizes and strides in the patch extraction process. The set of memory banks is therefore replicated for each scale, and $G(x)$ is split at multiple scales accordingly. Our loss function is given by the sum of the values from Eq.~\ref{eq:cx} computed at each scale, as follows:
\begin{equation}
\label{eq:cxms}
\mathcal{L}_{CXMS}(\bm{K}, \bm{B}) = \sum_s\mathcal{L}_{CX}^s(\bm{K}, \bm{B})
\end{equation}
where each scale $s$ implies a specific patch size and stride.

\subsection{Unpaired image-to-image translation baseline}
Our objective assumes the availability of a generated image $G(x)$ which is, in our task, the representation of a painting in the photo-realistic domain. In our work, we adopt a cycle-consistent adversarial framework~\cite{zhu2017unpaired} between the domain of paintings from a specific artist $X$ and the domain of realistic images $Y$. The data distributions are $x \sim p_{data}(x)$ and $y \sim p_{data}(y)$, while $G: X \rightarrow Y$ and $F: Y \rightarrow X$ are the mapping functions between the two domains. The two discriminators are denoted as $D_Y$ and $D_X$. The full cycle-consistent adversarial loss~\cite{zhu2017unpaired} is the following:
\begin{equation}
\label{eq:cycfull}
\begin{split}
\mathcal{L} _{cca}(G, F, D_X, D_Y)  &= \mathcal{L} _{GAN}(G,D_Y,X,Y)\\
                                    &+ \mathcal{L} _{GAN}(F,D_X,Y,X)\\
									&+ \mathcal{L} _{cyc}(G, F)
\end{split}
\end{equation}
where the two adversarial losses are:
\begin{equation}
\label{eq:adv1}
\begin{split}
\mathcal{L} _{GAN}(G,D_Y,X,Y)   &= \mathbb{E}_{y \sim p_{data}(y)}[logD_Y(y)]\\
                                &+ \mathbb{E}_{x \sim p_{data}(x)}[log(1 - D_Y(G(x)))]
\end{split}
\end{equation}
\begin{equation}
\label{eq:adv2}
\begin{split}
\mathcal{L} _{GAN}(F,D_X,Y,X)   &= \mathbb{E}_{x \sim p_{data}(x)}[logD_X(x)]\\
                                &+ \mathbb{E}_{y \sim p_{data}(y)}[log(1 - D_X(F(y)))]
\end{split}
\end{equation}
and the cycle consistency loss, which requires the original images $x$ and $y$ to be the same as the reconstructed ones, $F(G(x))$ and $G(F(y))$ respectively, is:
\begin{equation}
\label{eq:cyc}
\begin{split}
\mathcal{L} _{cyc}(G, F)&= \mathbb{E}_{x \sim p_{data}(x)}[\left\lVert F(G(x))-x \right\rVert]\\
                        &+ \mathbb{E}_{y \sim p_{data}(y)}[\left\lVert G(F(y))-y \right\rVert].
\end{split}
\end{equation}

\subsection{Full objective}
Our full semantically-aware translation loss is given by the sum of the baseline objective, \ie~Eq.~\ref{eq:cycfull}, and our patch-level similarity loss, \ie~Eq.~\ref{eq:cxms}:
\begin{equation}
\label{eq:fullobj}
\begin{split}
\mathcal{L}(G, F, D_X, D_Y, \bm{K}, \bm{B})   &=\mathcal{L}_{cca}(G, F, D_X, D_Y)\\ 
                                    &+ \lambda\mathcal{L}_{CXMS}(\bm{K}, \bm{B})
\end{split}
\end{equation}
where $\lambda$ controls our multi-scale contextual loss weight with respect to the baseline objective.


\section{Experimental results}
\label{sec:experiments}

\begin{table*}[t]
\small
\centering
\begin{tabular}{lcccccccc}
\toprule 
Method & Monet & Cezanne & Van Gogh & Ukiyo-e & Landscapes & Portraits & & Mean  \\
\midrule
Original paintings                      & 69.14 & 169.43 & 159.82 & 177.52 & 59.07 & 72.95 & & 117.99  \\
Style-transferred reals                 & 74.43 & 114.39 & 137.06 & 147.94 & 70.25 & 62.35 & & 101.07  \\
DRIT~\cite{DRIT}                        & 68.32 & 109.36 & 108.92 & 117.07 & 59.84 & 44.33 & & 84.64  \\
UNIT~\cite{liu2017unsupervised}         & 56.18 & 97.91 & 98.12 & 89.15 & 47.87 & 43.47 & & 72.12   \\
Cycle-GAN~\cite{zhu2017unpaired}        & 49.70 & 85.11 & 85.10 & 98.13 & 44.79 & \textbf{30.60} & & 65.57  \\
\midrule
\textbf{\ours}                          & \textbf{44.71} & \textbf{68.00} & \textbf{78.60} & \textbf{80.48} & \textbf{35.03} & 34.03 & & \textbf{56.81}  \\
\bottomrule
\end{tabular}
\caption{Evaluation in terms of Fr\'echet Inception Distance~\cite{heusel2017gans}.}
\label{tab:frechet_inception_distance}
\end{table*}

\begin{figure*}[t]
\centering
\small
\setlength{\tabcolsep}{.15em}
\begin{tabular}{cccc:ccccc}
& & \textbf{\ours} & & & Cycle-GAN~\cite{zhu2017unpaired} & UNIT~\cite{liu2017unsupervised} & DRIT~\cite{DRIT}  & Style-transferred reals  \\
\rotatebox{90}{\parbox[t]{0.95in}{\hspace*{\fill}Landscapes\hspace*{\fill}}} & & 
\includegraphics[width=0.185\linewidth]{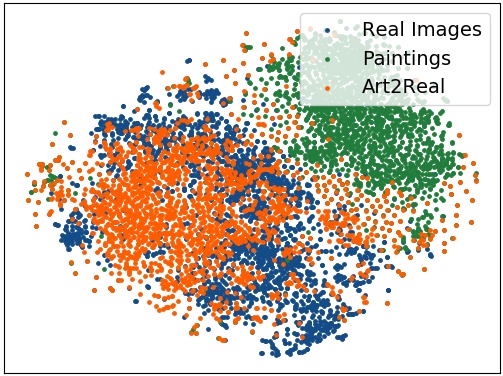} & & & 
\includegraphics[width=0.185\linewidth]{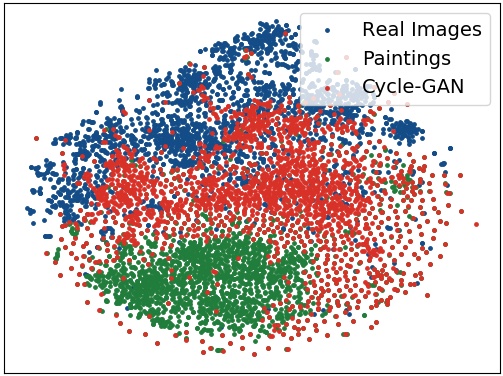}&
\includegraphics[width=0.185\linewidth]{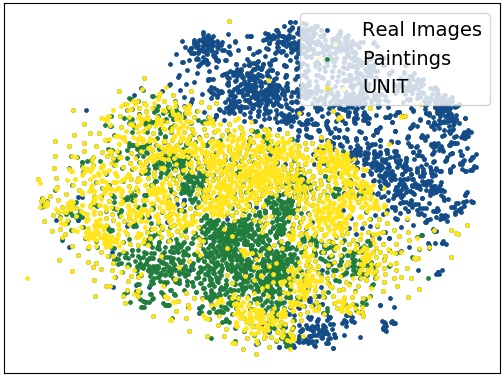} &
\includegraphics[width=0.185\linewidth]{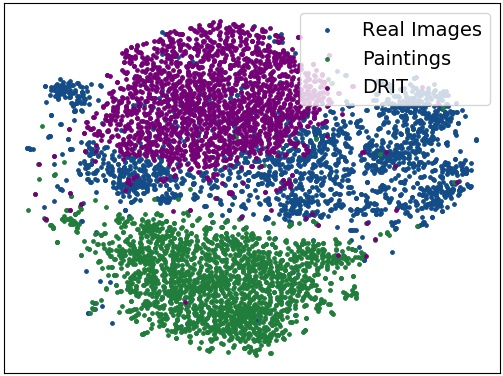} &
\includegraphics[width=0.185\linewidth]{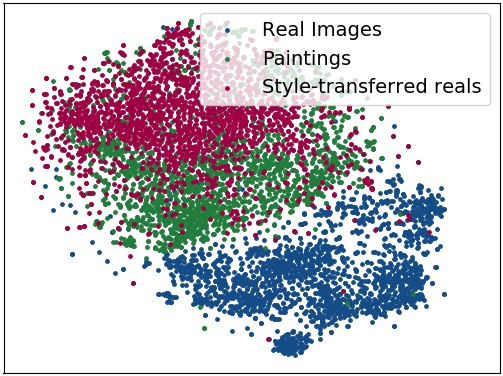} \\
\rotatebox{90}{\parbox[t]{0.95in}{\hspace*{\fill}Portraits\hspace*{\fill}}} & & 
\includegraphics[width=0.185\linewidth]{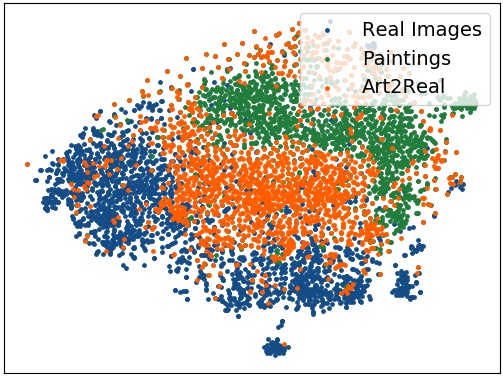} & & & 
\includegraphics[width=0.185\linewidth]{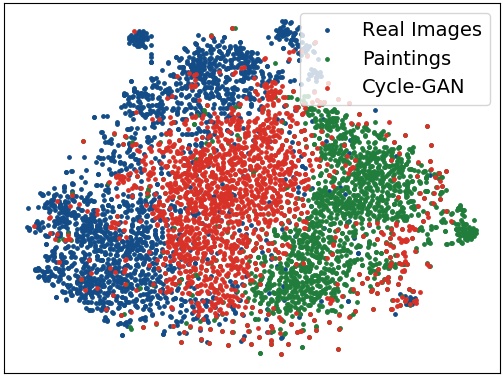}&
\includegraphics[width=0.185\linewidth]{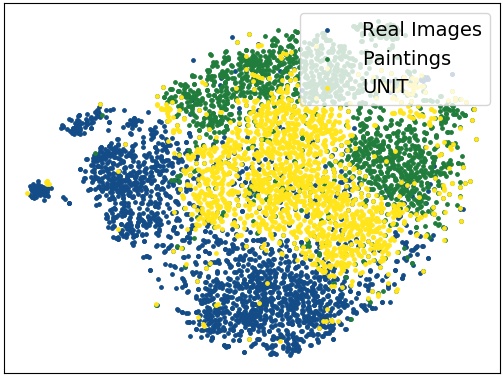} &
\includegraphics[width=0.185\linewidth]{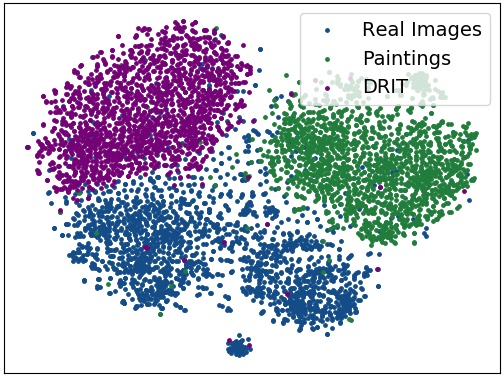} &
\includegraphics[width=0.185\linewidth]{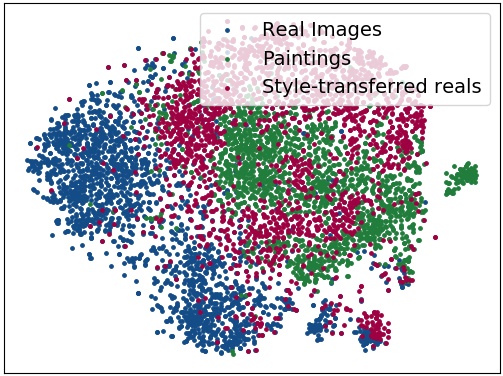} \\
\end{tabular}
\caption{Distribution of ResNet-152 features extracted from landscape and portrait images. Each row shows the results of our method and competitors on a specific setting.}
\label{fig:tsne}
\vspace{-.3cm}
\end{figure*}

\tinytit{Datasets}
In order to evaluate our approach, different sets of images, both from artistic and realistic domains, are used. Our tests involve both sets of paintings from specific artists and sets of artworks representing a given subject from different authors. We use paintings from Monet, Cezanne, Van Gogh, Ukiyo-e style and landscapes from different artists along with real photos of landscapes, keeping an underlying relationship between artistic and realistic domains. We also show results using portraits and real people photos. All artworks are taken from Wikiart.org, while landscape photos are downloaded from Flickr through the combination of tags \texttt{landscape} and \texttt{landscapephotography}. To obtain people photos, images are extracted from the CelebA dataset~\cite{liu2015faceattributes}. All the images are scaled to $256\times256$ pixels, and only RGB pictures are used. The size of each training set is, respectively, Monet: 1072, Cezanne: 583, Van Gogh: 400, Ukiyo-e: 825, landscape paintings: 2044, portraits: 1714, real landscape photographs: 2048, real people photographs: 2048. 

\tit{Architecture and training details}
To build generators and discriminators, we adapt generative networks from Johnson \etal~\cite{johnson2016perceptual}, with two stride-2 convolutions to downsample the input, several residual blocks and two stride-1/2 convolutional layers for upsampling. Discriminative networks are PatchGANs~\cite{isola2017image,ledig2017photo,li2016precomputed} which classify each square patch of an image as real or fake. 

Memory banks of real patches are built using all the available real images, \ie~2048 images both for landscapes and for people faces, and are kept constant during training. Masks of the paintings, after epoch 40, are regularly updated every 20 epochs with those from the generated images. Patches are extracted at three different scales: $4\times 4$, $8 \times 8$ and $16\times 16$, using three different stride values: 4, 5 and 6 respectively. The same patch sizes and strides are adopted when splitting the generated image, in order to compute affinities and the contextual loss. We use a multi-scale contextual loss weight $\lambda$, in Eq. \ref{eq:fullobj}, equal to $0.1$.

We train the model for $300$ epochs through the Adam optimizer~\cite{kingma2014adam} and using mini-batches with a single sample. A learning rate of $0.0002$ is kept constant for the first $100$ epochs, making it linearly decay to zero over the next $200$ epochs. An early stopping technique is used to reduce training times. In particular, at each epoch the Fr\'echet Inception Distance (FID)~\cite{heusel2017gans} is computed between our generated images and the set of real photos: if it does not decrease for 30 consecutive epochs, the training is stopped. We initialize the weights of the model from a Gaussian distribution with 0 mean and standard deviation $0.02$.

\tit{Competitors}
To compare our results with those from state of the art techniques, we train Cycle-GAN~\cite{zhu2017unpaired}, UNIT~\cite{liu2017unsupervised} and DRIT~\cite{DRIT} approaches on the previously described datasets. The adopted code comes from the authors' implementations and can be found in their GitHub repositories. The number of epochs and other training parameters are those suggested by the authors, except for DRIT~\cite{DRIT}: to enhance the quality of the results generated by this competitor, after contacting the authors we employed spectral normalization and manually chose the best epoch through visual inspection and by computing the FID~\cite{heusel2017gans} measure. Moreover, being DRIT~\cite{DRIT} a diverse image-to-image translation framework, its performance depends on the choice of an attribute from the attribute space of the realistic domain. For fairness of comparison, we generate a single realistic image using a randomly sampled attribute. We also show quantitative results of applying the style transfer approach from Gatys \etal~\cite{gatys2015neural}, with content images taken from the realistic datasets and style images randomly sampled from the paintings, for each set.

\subsection{Visual quality evaluation}
We evaluate the visual quality of our generated images using both automatic evaluation metrics and user studies.

\tit{Fr\'echet Inception Distance} To numerically assess the quality of our generated images, we employ the Fr\'echet Inception Distance~\cite{heusel2017gans}. It measures the
difference of two Gaussians, and it is also known as Wasserstein-2 distance~\cite{vaserstein1969markov}. The FID $d$ between a Gaussian $G_1$ with mean and covariance $(m_1, C_1)$ and a Gaussian $G_2$ with mean and covariance $(m_2,C_2)$ is given by:
\begin{equation}
\label{eq:fid}
d^2(G_1,G_2) = \left\lVert m_1-m_2\right\rVert_2^2 + Tr(C_1+C_2-2(C_1C_2)^{1/2})
\end{equation}
For our evaluation purposes, the two Gaussians are fitted on Inception-v3~\cite{szegedy2016rethinking} activations of real and generated images, respectively. The lower the Fr\'echet Inception Distance between these Gaussians, the more generated and real data distributions overlap, \ie~the realism of generated images increases when the FID decreases. Table \ref{tab:frechet_inception_distance} shows FID values for our model and a number of competitors. As it can be observed, the proposed approach produces a lower FID on all settings, except for portraits, in which we rank second after Cycle-GAN. Results thus confirm the capabilities of our method in producing images which looks realistic to pre-trained CNNs.

\begin{table}[t]
\small
\centering
\setlength{\tabcolsep}{.3em}
\begin{tabular}{lccc}
\toprule
 & Cycle-GAN~\cite{zhu2017unpaired} & UNIT~\cite{liu2017unsupervised}  & DRIT~\cite{DRIT} \\
\midrule
Realism             & 36.5\% & 27.9\% & 14.2\%\\
\midrule
Coherence           & 48.4\% & 25.5\% & 7.3\%\\
\bottomrule
\end{tabular}
\caption{User study results. We report the percentage of times an image from a competitor was preferred against ours. Our method is always preferred more than 50\% of the times.}
\label{tab:user_study}
\vspace{-.3cm}
\end{table}

\begin{figure*}[tb]
\centering
\small
\setlength{\tabcolsep}{.18em}
\begin{tabular}{cc:ccccc}
Original Painting & & & \textbf{\ours} & Cycle-GAN~\cite{zhu2017unpaired} & UNIT~\cite{liu2017unsupervised} & DRIT~\cite{DRIT}  \\
\includegraphics[width=0.185\linewidth]{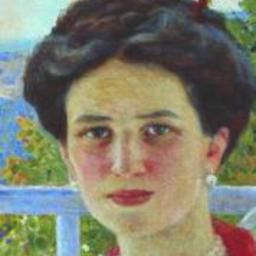}& & &
\includegraphics[width=0.185\linewidth]{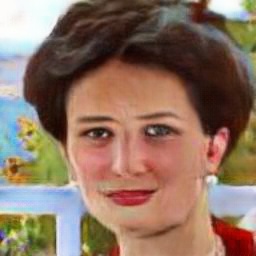} &
\includegraphics[width=0.185\linewidth]{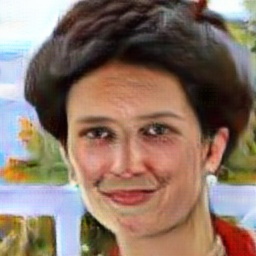} &
\includegraphics[width=0.185\linewidth]{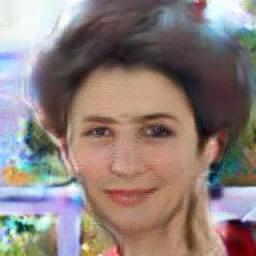} &
\includegraphics[width=0.185\linewidth]{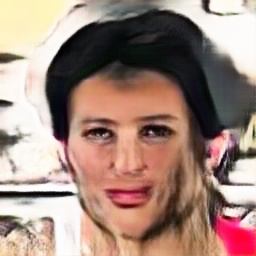} \\
\includegraphics[width=0.185\linewidth]{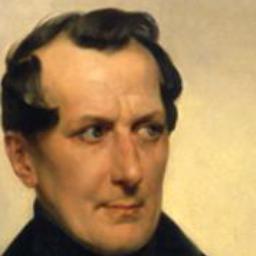}& & &
\includegraphics[width=0.185\linewidth]{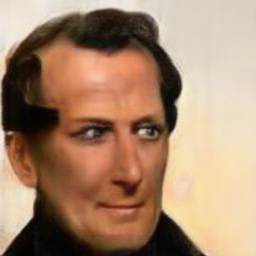} &
\includegraphics[width=0.185\linewidth]{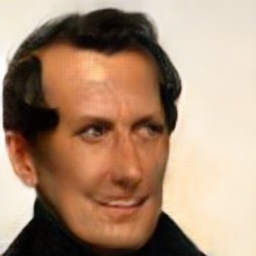} &
\includegraphics[width=0.185\linewidth]{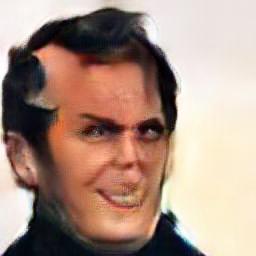} &
\includegraphics[width=0.185\linewidth]{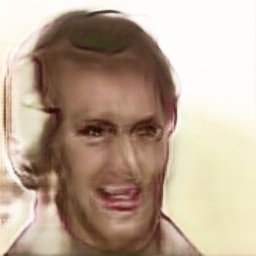} \\
\includegraphics[width=0.185\linewidth]{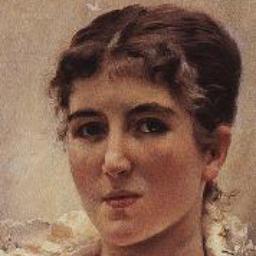}& & &
\includegraphics[width=0.185\linewidth]{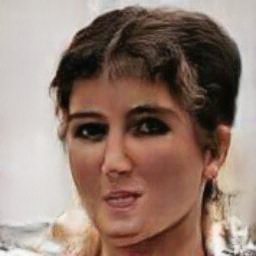} &
\includegraphics[width=0.185\linewidth]{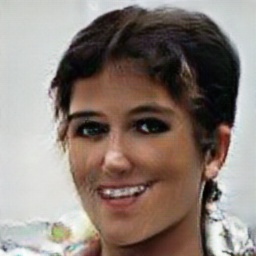} &
\includegraphics[width=0.185\linewidth]{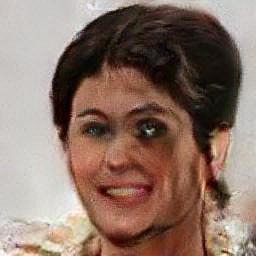} &
\includegraphics[width=0.185\linewidth]{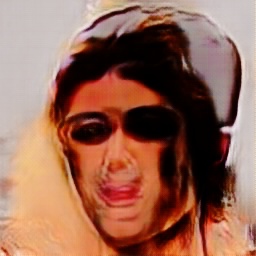} \\
\end{tabular}
\caption{Qualitative results on portraits. Our method can preserve facial expressions and reduce the amount of artifacts with respect to Cycle-GAN~\cite{zhu2017unpaired}, UNIT~\cite{liu2017unsupervised}, and DRIT~\cite{DRIT}.}
\label{fig:results_2}
\vspace{-.3cm}
\end{figure*}

\begin{figure*}[tb]
\centering
\small
\setlength{\tabcolsep}{.18em}
\begin{tabular}{cc:ccccc}
Original Painting & & & \textbf{\ours} & Cycle-GAN~\cite{zhu2017unpaired} & UNIT~\cite{liu2017unsupervised} & DRIT~\cite{DRIT}  \\
\includegraphics[width=0.185\linewidth]{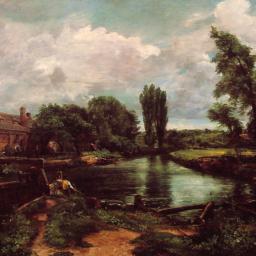}& & &
\includegraphics[width=0.185\linewidth]{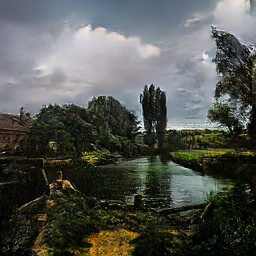} &
\includegraphics[width=0.185\linewidth]{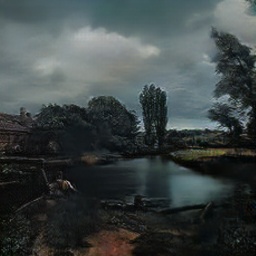} &
\includegraphics[width=0.185\linewidth]{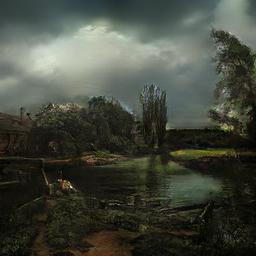} &
\includegraphics[width=0.185\linewidth]{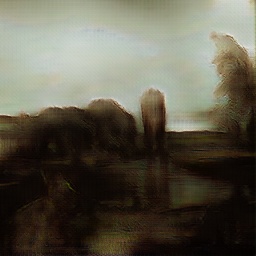} \\
\includegraphics[width=0.185\linewidth]{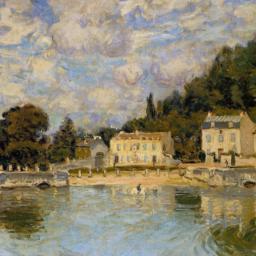}& & &
\includegraphics[width=0.185\linewidth]{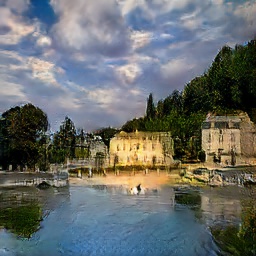} &
\includegraphics[width=0.185\linewidth]{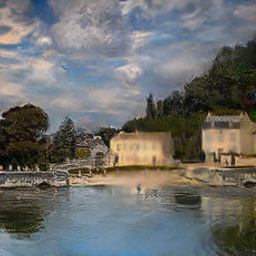} &
\includegraphics[width=0.185\linewidth]{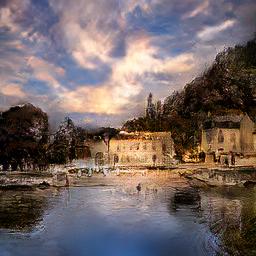} &
\includegraphics[width=0.185\linewidth]{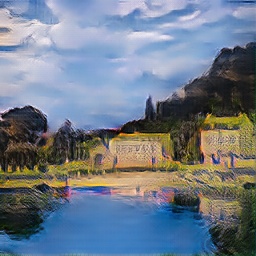} \\
\includegraphics[width=0.185\linewidth]{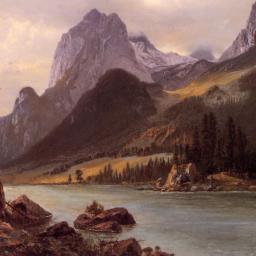}& & &
\includegraphics[width=0.185\linewidth]{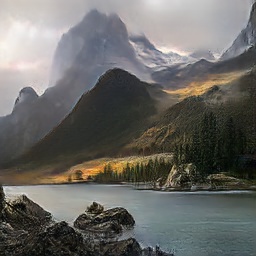} &
\includegraphics[width=0.185\linewidth]{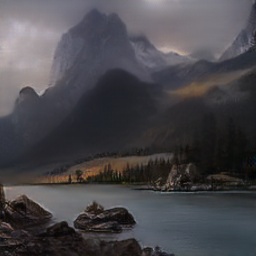} &
\includegraphics[width=0.185\linewidth]{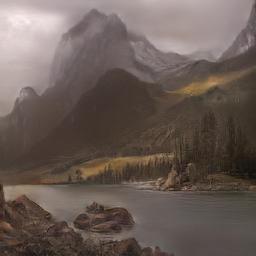} &
\includegraphics[width=0.185\linewidth]{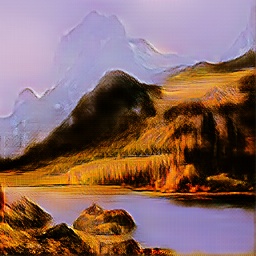} \\
\end{tabular}
\caption{Qualitative results on landscape paintings. Results generated by our approach show increased realism and reduced blur when compared with those from Cycle-GAN~\cite{zhu2017unpaired}, UNIT~\cite{liu2017unsupervised}, and DRIT~\cite{DRIT}.}
\label{fig:results}
\end{figure*}

\tit{Human judgment}
In order to evaluate the visual quality of our generated images, we conducted a user study on the Figure Eight crowd-sourcing platform. In particular, we assessed both the realism of our results and their coherence with the original painting. To this aim, we conducted two different evaluation processes, which are detailed as follows:
\begin{itemize}[noitemsep,topsep=0pt]
\item In the \textit{Realism} evaluation, we asked the user to select the most realistic image between the two shown, both obtained from the same painting, one from our method and the other from a competitor;
\item In the \textit{Coherence} evaluation, we presented the user the original painting and two generated images which originate from it, asking to select the most faithful to the artwork. Again, generated images come from our method and a competitor.
\end{itemize} 
Each test involved our method and one competitor at a time leading to six different tests, considering three competitors: Cycle-GAN~\cite{zhu2017unpaired}, UNIT~\cite{liu2017unsupervised}, and DRIT~\cite{DRIT}. A set of 650 images were randomly sampled for each test, and each image pair was evaluated from three different users. Each user, to start the test, was asked to successfully evaluate eight example pairs where one of the two images was definitely better than the other. A total of 685 evaluators were involved in our tests. Results are presented in Table~\ref{tab:user_study}, showing that our generated images are always chosen more than $50\%$ of the times.

\begin{table}[t]
\small
\centering
\setlength{\tabcolsep}{.3em}
\resizebox{\linewidth}{!}{
\begin{tabular}{lccc}
\toprule
Method & Classification & Segmentation & Detection \\
\midrule
Real Photos                             & 3.99 & 0.63 & 2.03 \\
\midrule
Original paintings                      & 4.81 & 0.67 & 2.58 \\
Style-transferred reals                 & 5.39 & 0.70 & 2.89 \\
DRIT~\cite{DRIT}                        & 5.14 & 0.67 & 2.56 \\
UNIT~\cite{liu2017unsupervised}         & 4.88 & 0.69 & 2.54 \\
Cycle-GAN~\cite{zhu2017unpaired}        & 4.81 & 0.67 & 2.50 \\
\midrule
\textbf{\ours}                          & \textbf{4.50} & \textbf{0.66} & \textbf{2.42}\\
\bottomrule
\end{tabular}
}
\caption{Mean entropy values for classification, segmentation, and detection of images generated through our method and through competitor methods.}
\label{tab:entropy}
\vspace{-.3cm}
\end{table}

\subsection{Reducing the domain shift}
\label{sub:reduce_shift}
We evaluate the capabilities of our model to reduce the domain shift between artistic and real data, by analyzing the performance of pre-trained convolutional models and visualizing the distributions of CNN features.

\tinytit{Entropy analysis}
Pre-trained architectures show increased performances on images synthesized by our approach, in comparison with original paintings and images generated by other approaches. We visualize this by computing the entropy of the output of state of the art architectures: the lower the entropy, the lower the uncertainty of the model about its result. We evaluate the entropy on classification, semantic segmentation, and detection tasks, adopting a ResNet-152~\cite{he2016deep}
trained on ImageNet~\cite{imagenet_cvpr09}, Hu~\etal~\cite{hu2018learning}'s model and Faster R-CNN~\cite{ren2017faster} trained on the Visual Genome~\cite{krishnavisualgenome,Anderson2017up-down}, respectively. Table~\ref{tab:entropy} shows the average image entropy for classification, the average pixel entropy for segmentation and the average bounding-box entropy for detection, computed on all the artistic, realistic and generated images available. Our approach is able to generate images which lower the entropy, on average, for each considered task with respect to paintings and images generated by the competitors.

\tit{Feature distributions visualization} To further validate the domain shift reduction between real images and generated ones, we visualize the distributions of features extracted from a CNN. In particular, for each image, we extract a visual feature vector coming from the average pooling layer of a ResNet-152~\cite{he2016deep}, and we project it into a 2-dimensional space by using the t-SNE algorithm~\cite{maaten2008visualizing}. Fig.~\ref{fig:tsne} shows the feature distributions on two different sets of paintings (\ie, landscapes and portraits) comparing our results with those of competitors. Each plot represents the distribution of visual features extracted from paintings belonging to a specific set, from the corresponding images generated by our model or by one of the competitors, and from the real photographs depicting landscapes or, in the case of portraits, human faces. As it can be seen, the distributions of our generated images are in general closer to the distributions of real images than to those of paintings, thus confirming the effectiveness of our model in the domain shift reduction.

\subsection{Qualitative results}
Besides showing numerical improvements with respect to state of the art approaches, we present some qualitative results coming from our method, compared to those from Cycle-GAN~\cite{zhu2017unpaired}, UNIT~\cite{liu2017unsupervised}, and DRIT~\cite{DRIT}. We show examples of landscape and portrait translations in Fig.~\ref{fig:results_2} and ~\ref{fig:results}. Many other samples from all settings can be found in the Supplementary material. We observe increased realism in our generated images, due to more detailed elements and fewer blurred areas, especially in the landscape results. Portrait samples reveal that brush strokes disappear completely, leading to a photo-realistic visualization. Our results contain fewer artifacts and are more faithful to the paintings, more often preserving the original facial expression.

\section{Conclusion}
\label{sec:conclusion}
We have presented \textit{\ours}, an approach to translate paintings to photo-realistic visualizations. Our research is motivated by the need of reducing the domain gap between artistic and real data, which prevents the application of recent techniques to art. The proposed approach generates realistic images by copying from sets of real images, in a semantically aware manner and through efficient memory banks. This is paired with an image-to-image translation architecture, which ultimately leads to the final result. Quantitative and qualitative evaluations, conducted on artworks of different artists and styles, have shown the effectiveness of our method in comparison with image-to-image translation algorithms. Finally, we also showed how generated images can enhance the performance of pre-trained architectures.

{\small
\bibliographystyle{ieee}
\bibliography{egbib}
}

\newpage 
\appendix
\section*{Supplementary material}
In the following, we present additional material about our method. In particular, we provide a visualization of the segmentation maps, an analysis of the importance of multi-scale, and additional quantitative and qualitative results.

\section{Segmentation maps}
\begin{figure}[t]
    \centering
    \includegraphics[width=0.98\linewidth]{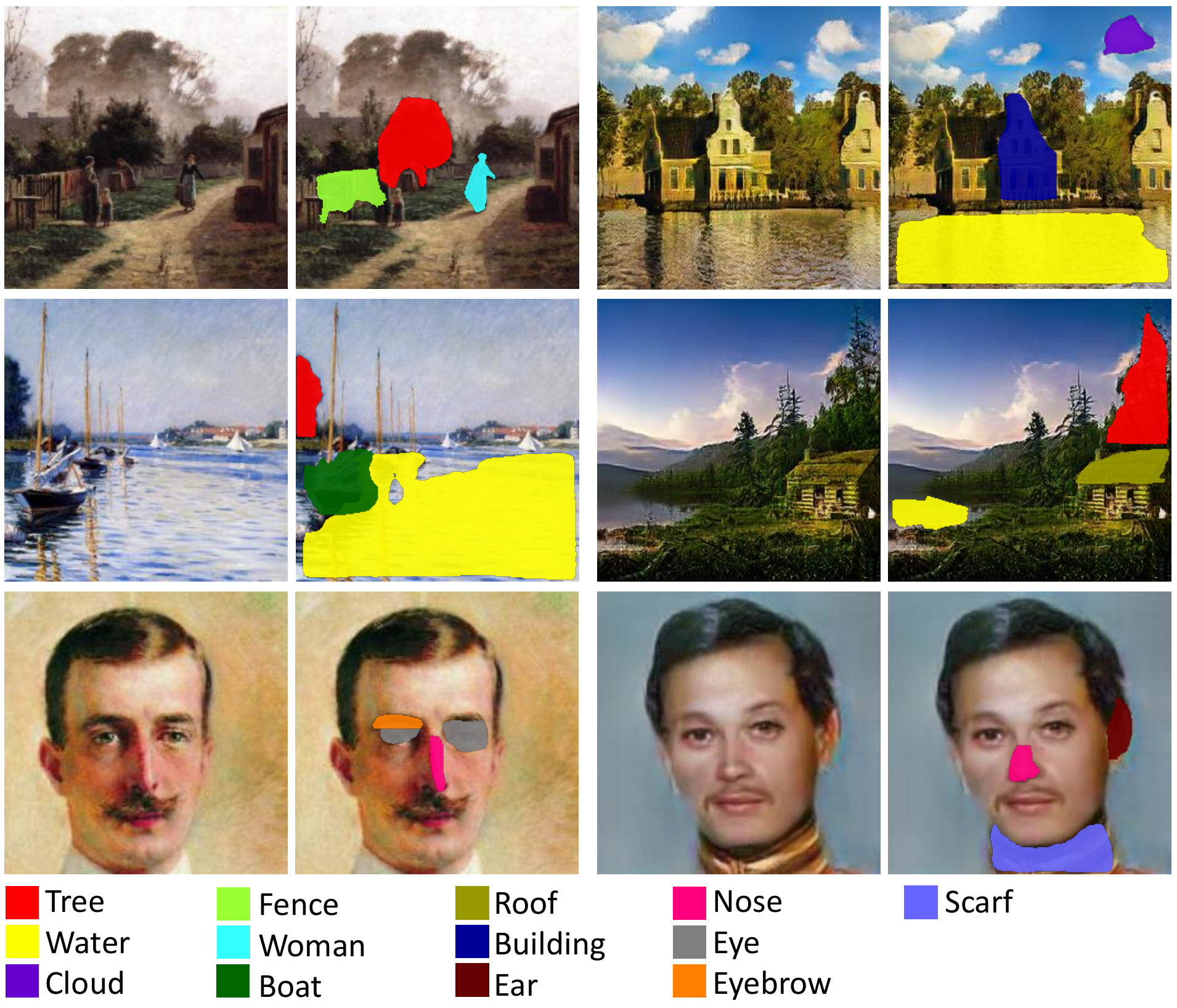}
    \caption{Segmentation masks visualization. The first two columns show original paintings, while the third and the fourth columns show generated images. Only some of the extracted masks are visible.}
    \label{fig:seg_masks}
\end{figure}

In Fig.~\ref{fig:seg_masks}, we show some qualitative examples of segmentation masks extracted on paintings and generated images, through the model from Hu~\etal~\cite{hu2018learning}. Each color represents a specific class label. Only the most relevant masks are shown for each image. It can be observed that the segmentation strategy extracts meaningful semantic regions from the input images, thus enforcing the retrieval of semantically correct patches on large portions of the image.  Overall, we found that the use of semantic segmentation can greatly improve results. While some image regions might not be labelled, a sufficiently realistic appearance can still be recovered from the background memory bank.

\section{Multi-scale importance}
In order to evaluate the contribution of the multi-scale approach to the realism of the generation, we run a set of experiments without the multi-scale variant. We use a single scale, \ie~a patch size of 16 and a stride of 6, and train our model on Monet, landscape and portrait settings. The full objective, in this case, is that presented in Eq.~5 of the main paper. We then compare the FID~\cite{heusel2017gans} values obtained through our approach with and without the multi-scale variant. Results are presented in Table~\ref{tab:multi-scale}. As it can be seen, the multi-scale strategy effectively increases the realism of the generation, outperforming by a clear margin the single-scale baseline on almost all settings.

\section{Additional experimental results}
Here we present additional quantitative results, computing the FID~\cite{heusel2017gans} with different layers of Inception-v3~\cite{szegedy2016rethinking} and showing the distribution of ResNet-152~\cite{he2016deep} features extracted from all the available settings.

\tit{Fr\'echet Inception Distance} 
In the main paper, we showed how our model is able to generate images which lower the FID~\cite{heusel2017gans} with respect to real images, fitting the Gaussians on the final average pooling layer features of Inception-v3~\cite{szegedy2016rethinking} (2048-d). In Table~\ref{tab:frechet_inception_distance_supp}, we also show FID values obtained fitting the two Gaussians on the pre-auxiliary classifier layer features (768-d) and on the second max-pooling layer features (192-d). The FID value is computed for our model and for a number of competitors and again our model produces a lower FID on almost all the settings. Note that FID values computed at different layers have different magnitude and are not directly comparable.

\begin{table}[t]
\small
\centering
\setlength{\tabcolsep}{.3em}
\resizebox{\linewidth}{!}{
\begin{tabular}{lccccc}
\toprule 
Method & Monet & Landscapes & Portraits & & Mean  \\
\midrule
Single scale        & 46.28 & 35.88 & 34.74 & & 38.97 \\
\midrule
\textbf{\ours} (multi-scale)    & \textbf{44.71} & \textbf{35.03} & \textbf{34.03} & & \textbf{37.92} \\
\bottomrule
\end{tabular}
}
\caption{Multi-scale importance analysis in terms of Fr\'echet Inception Distance~\cite{heusel2017gans}.}
\label{tab:multi-scale}
\vspace{-0.3cm}
\end{table}

\begin{table*}[t]
\small
\centering
\begin{tabular}{lcccccccc}
\toprule 
Method & Monet & Cezanne & Van Gogh & Ukiyo-e & Landscapes & Portraits & & Mean  \\
\midrule
& \multicolumn{8}{c}{\cellcolor{Gray}768 dimensions} \\
\midrule
Original paintings                      & 0.45 & 0.94 & 1.03 & 1.34 & 0.37 & 0.42 & & 0.76 \\
Style-transferred reals                 & 0.58 & 0.94 & 1.12 & 1.23 & 0.56 & 0.36 & & 0.80 \\
DRIT~\cite{DRIT}                        & 0.41 & 0.54 & 0.56 & 0.60 & 0.37 & 0.28 & & 0.46 \\
UNIT~\cite{liu2017unsupervised}         & 0.30 & 0.43 & 0.44 & 0.35 & 0.25 & 0.25 & & 0.34 \\
Cycle-GAN~\cite{zhu2017unpaired}        & 0.29 & 0.37 & 0.36 & 0.43 & 0.24 & \textbf{0.16} & & 0.31 \\
\midrule
\textbf{\ours}                                    & \textbf{0.21} & \textbf{0.30} & \textbf{0.35} & \textbf{0.31} & \textbf{0.17} & 0.19 & & \textbf{0.26} \\
\toprule
& \multicolumn{8}{c}{\cellcolor{Gray}192 dimensions} \\
\midrule
Original paintings                      & 0.95 & 1.67 & 3.96 & 1.86 & 0.49 & 0.22 & & 1.53  \\
Style-transferred reals                 & 0.97 & 1.76 & 4.09 & 2.44 & 0.55 & 0.21 & & 1.67 \\
DRIT~\cite{DRIT}                        & 0.30 & 0.33 & 0.40 & 0.38 & 0.49 & 0.11 & & 0.34 \\
UNIT~\cite{liu2017unsupervised}         & 0.26 & 0.26 & 0.37 & \textbf{0.16} & 0.21 & 0.07 & & 0.22 \\
Cycle-GAN~\cite{zhu2017unpaired}        & 0.26 & 0.31 & 0.18 & 0.19 & 0.55 & \textbf{0.03} & & 0.25 \\
\midrule
\textbf{\ours}                                  & \textbf{0.10} & \textbf{0.13} & \textbf{0.12} & 0.17 & \textbf{0.19} & 0.05 & & \textbf{0.13}\\
\bottomrule
\end{tabular}
\caption{Evaluation in terms of Fr\'echet Inception Distance~\cite{heusel2017gans}.}
\label{tab:frechet_inception_distance_supp}
\end{table*}

\begin{figure*}[tb]
\centering
\small
\setlength{\tabcolsep}{.15em}
\begin{tabular}{cccc:ccccc}
& & \textbf{\ours} & & & Cycle-GAN~\cite{zhu2017unpaired} & UNIT~\cite{liu2017unsupervised} & DRIT~\cite{DRIT}  & Style-transferred reals  \\
\rotatebox{90}{\parbox[t]{0.95in}{\hspace*{\fill}Monet\hspace*{\fill}}} & & 
\includegraphics[width=0.185\linewidth]{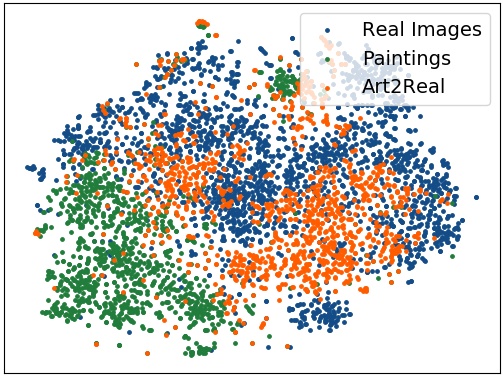}& & & 
\includegraphics[width=0.185\linewidth]{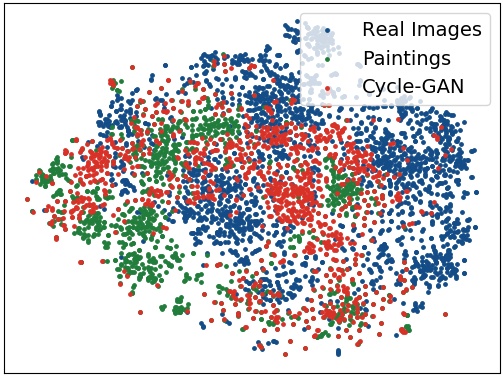}&
\includegraphics[width=0.185\linewidth]{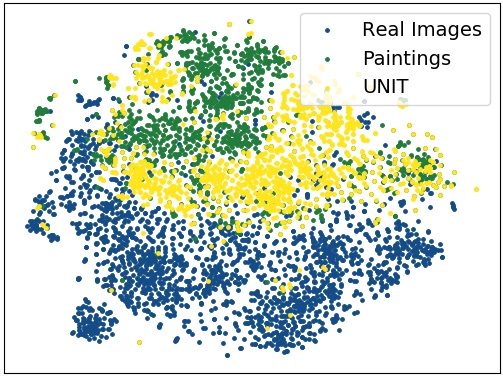} &
\includegraphics[width=0.185\linewidth]{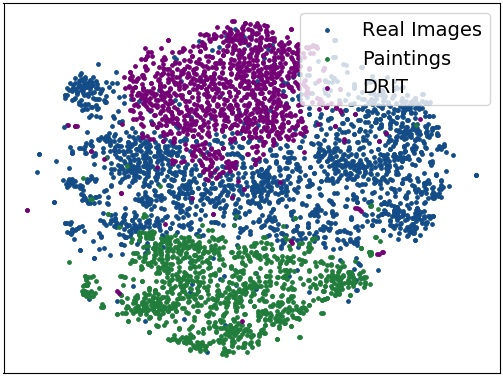} &
\includegraphics[width=0.185\linewidth]{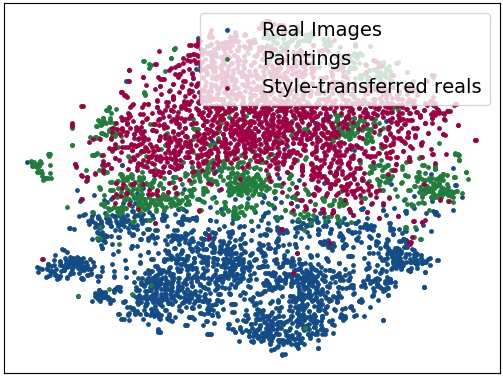} \\
\rotatebox{90}{\parbox[t]{0.95in}{\hspace*{\fill}Cezanne\hspace*{\fill}}} & & 
\includegraphics[width=0.185\linewidth]{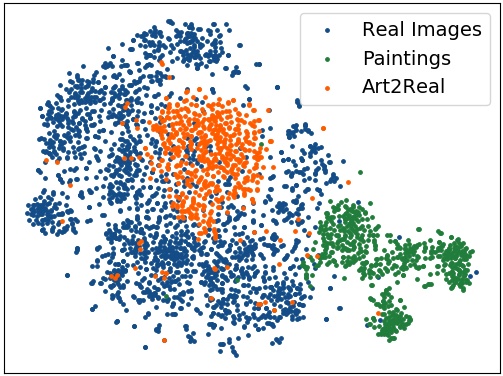}& & &
\includegraphics[width=0.185\linewidth]{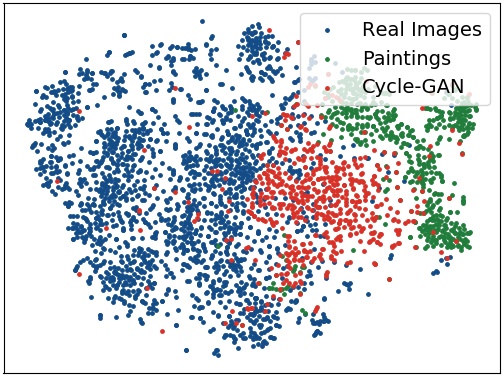}&
\includegraphics[width=0.185\linewidth]{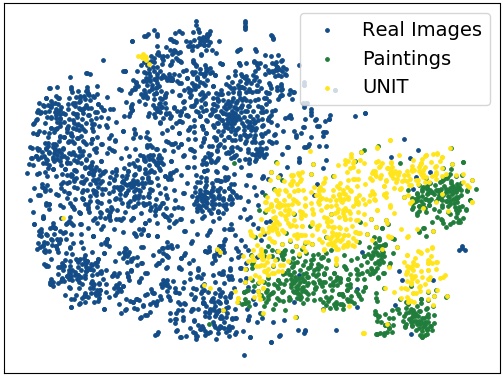} &
\includegraphics[width=0.185\linewidth]{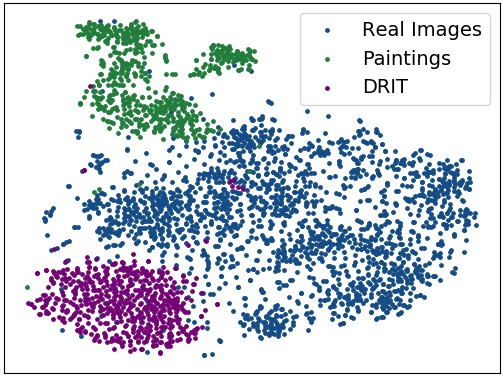} &
\includegraphics[width=0.185\linewidth]{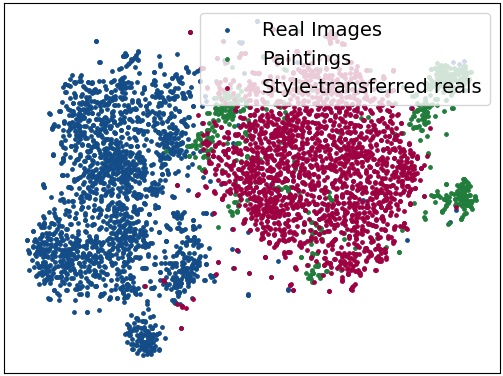} \\
\rotatebox{90}{\parbox[t]{0.95in}{\hspace*{\fill}Van Gogh\hspace*{\fill}}} & & 
\includegraphics[width=0.185\linewidth]{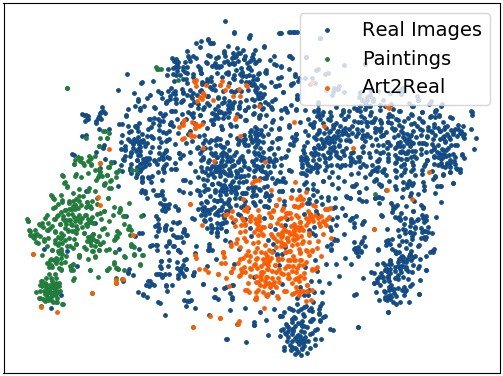}& & & 
\includegraphics[width=0.185\linewidth]{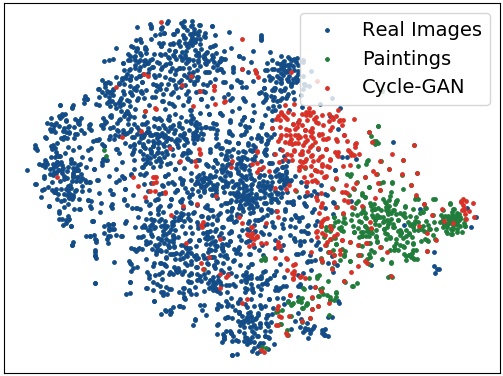}&
\includegraphics[width=0.185\linewidth]{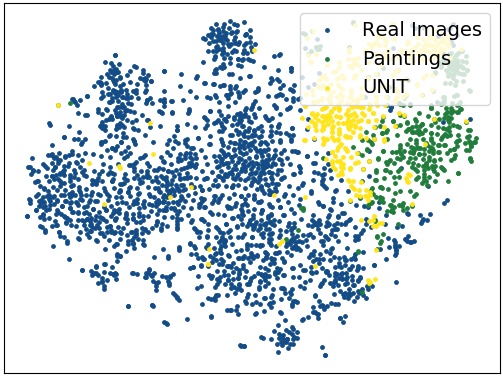} &
\includegraphics[width=0.185\linewidth]{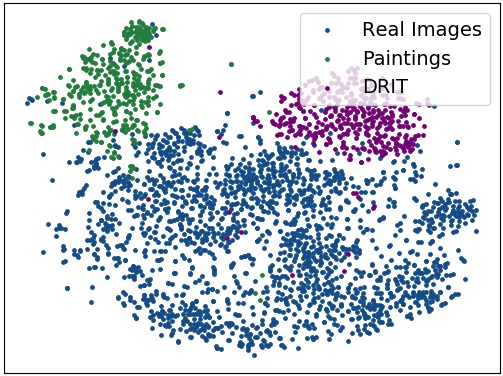} &
\includegraphics[width=0.185\linewidth]{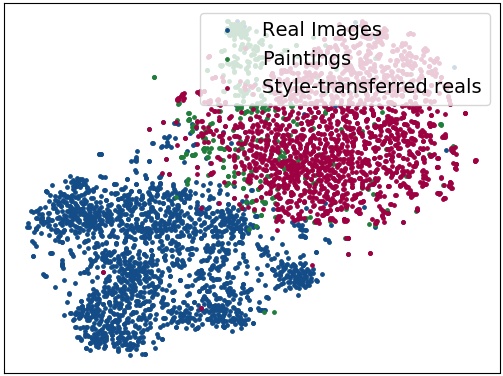} \\
\rotatebox{90}{\parbox[t]{0.95in}{\hspace*{\fill}Ukiyo-e\hspace*{\fill}}} & & 
\includegraphics[width=0.185\linewidth]{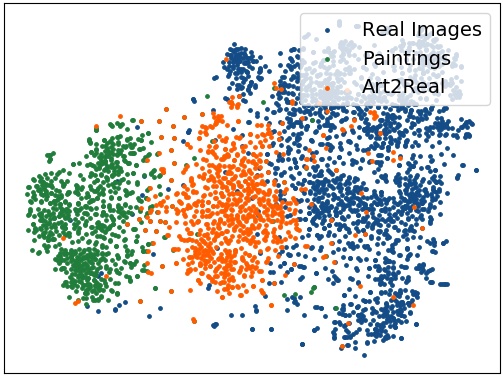} & & & 
\includegraphics[width=0.185\linewidth]{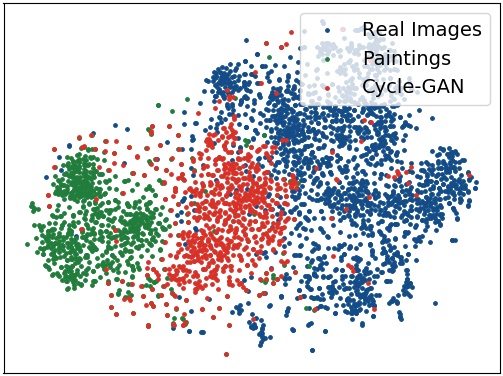}&
\includegraphics[width=0.185\linewidth]{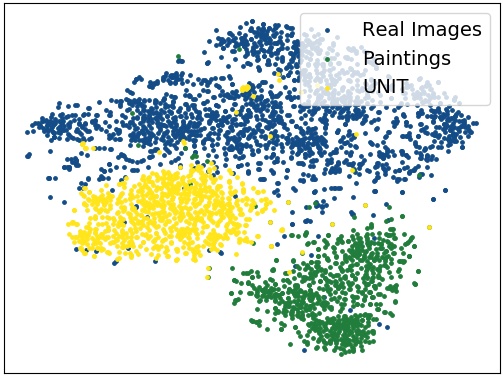} &
\includegraphics[width=0.185\linewidth]{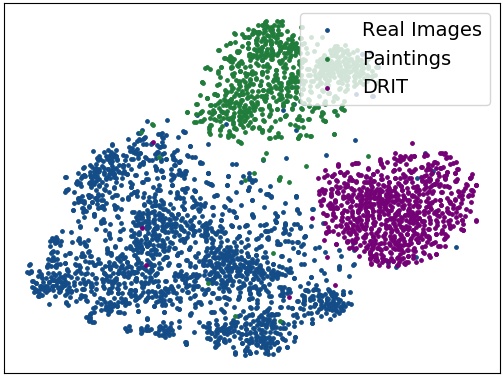} &
\includegraphics[width=0.185\linewidth]{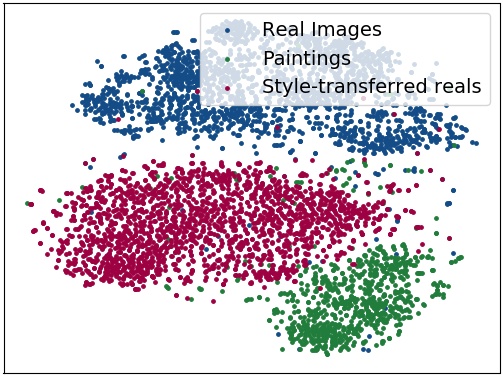} \\
\end{tabular}
\caption{Distribution of ResNet-152 features extracted from Monet, Cezanne, Van Gogh and Ukiyo-e images. Each row shows the results of our method and competitors on a specific setting.}
\label{fig:tsne_supp}
\vspace{-0.3cm}
\end{figure*}

\tit{Feature distributions visualization}
Fig.~\ref{fig:tsne_supp} shows the feature distribution visualizations of our method and competitors computed on Monet, Cezanne, Van Gogh and Ukiyo-e images. As previously mentioned, for each considered setting, we extract image features from the average pooling layer of a ResNet-152~\cite{he2016deep} and we use the t-SNE algorithm~\cite{maaten2008visualizing} to project them into a 2-dimensional space. Each plot reports the distributions of visual features extracted from real photos, original paintings and the corresponding translations generated by our model or by one of the competitors. Also for these settings, the distributions of visual features extracted from our generated images are very close to the distributions of real photos, thus further confirming a greater reduction of domain shift compared to that of competitors. 

\section{Additional qualitative results}

Several other qualitative results are shown in the rest of the supplementary. Firstly, we report sample images generated by our model taking as input sample paintings depicting landscapes and portraits. Secondly, we show additional qualitative comparisons with respect to Cycle-GAN~\cite{zhu2017unpaired}, UNIT~\cite{liu2017unsupervised}, and DRIT~\cite{DRIT} on all considered settings. Overall, the results demonstrate that our model is able to generate more realistic images, creating fewer artifacts and better preserving the original contents, facial expressions, and colors of original paintings.


\begin{figure*}[tb]
\centering
\small
\setlength{\tabcolsep}{.15em}
\begin{tabular}{cc:ccccccc:cc}
Original Painting & & & \textbf{\ours} & & & & Original Painting & & & \textbf{\ours} \\
\includegraphics[width=0.235\linewidth]{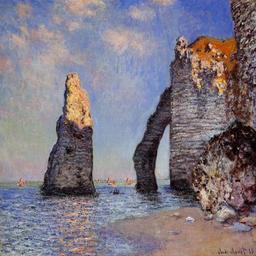}& & &
\includegraphics[width=0.235\linewidth]{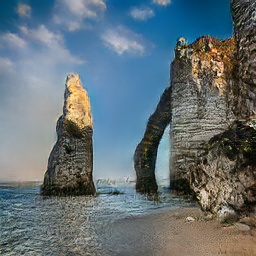} & & & &
\includegraphics[width=0.235\linewidth]{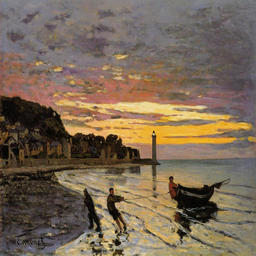} & & &
\includegraphics[width=0.235\linewidth]{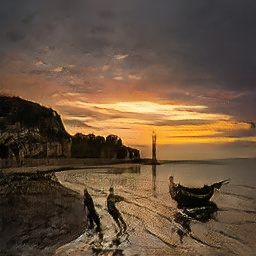} \\
\includegraphics[width=0.235\linewidth]{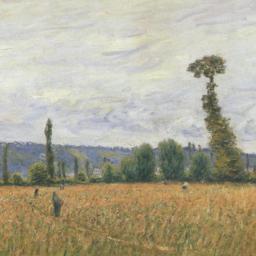}& & &
\includegraphics[width=0.235\linewidth]{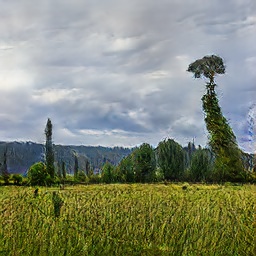} & & & &
\includegraphics[width=0.235\linewidth]{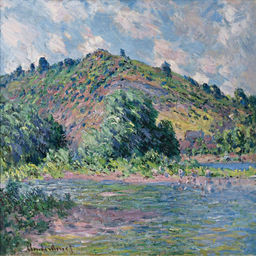} & & &
\includegraphics[width=0.235\linewidth]{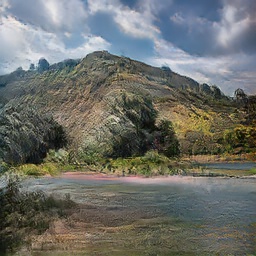} \\
\includegraphics[width=0.235\linewidth]{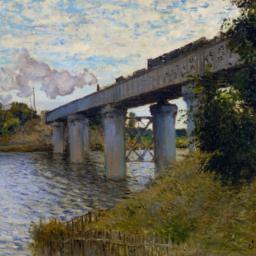}& & &
\includegraphics[width=0.235\linewidth]{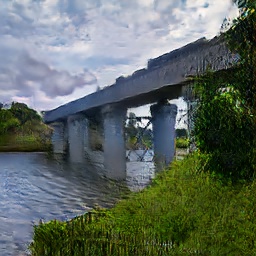} & & & &
\includegraphics[width=0.235\linewidth]{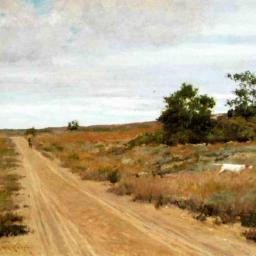} & & &
\includegraphics[width=0.235\linewidth]{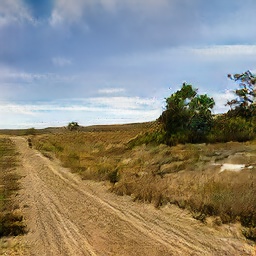} \\
\includegraphics[width=0.235\linewidth]{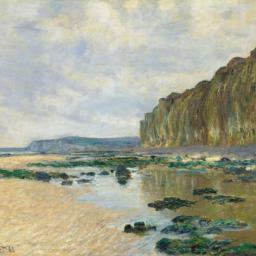}& & &
\includegraphics[width=0.235\linewidth]{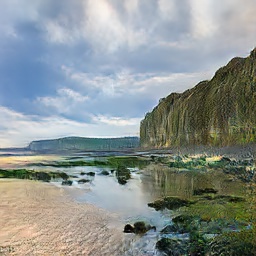} & & & &
\includegraphics[width=0.235\linewidth]{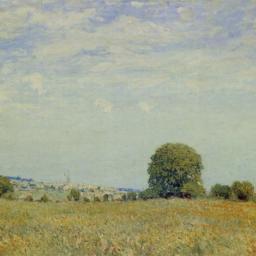} & & &
\includegraphics[width=0.235\linewidth]{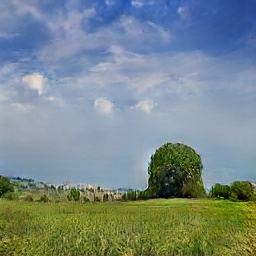} \\
\includegraphics[width=0.235\linewidth]{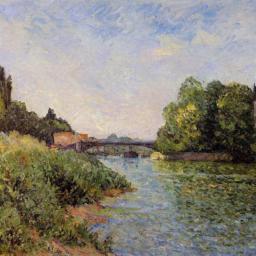}& & &
\includegraphics[width=0.235\linewidth]{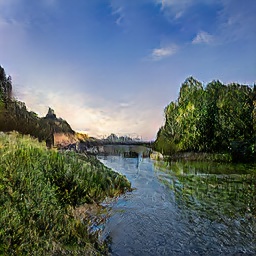} & & & &
\includegraphics[width=0.235\linewidth]{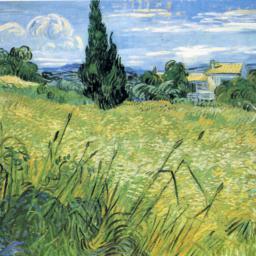} & & &
\includegraphics[width=0.235\linewidth]{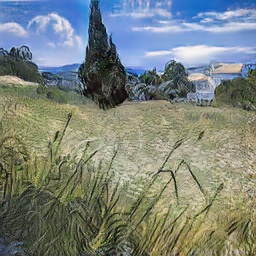} \\
\end{tabular}
\label{fig:our_results_supp1}
\end{figure*}

\begin{figure*}[tb]
\centering
\small
\setlength{\tabcolsep}{.15em}
\begin{tabular}{cc:ccccccc:cc}
Original Painting & & & \textbf{\ours} & & & & Original Painting & & & \textbf{\ours} \\
\includegraphics[width=0.235\linewidth]{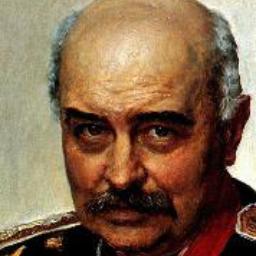}& & &
\includegraphics[width=0.235\linewidth]{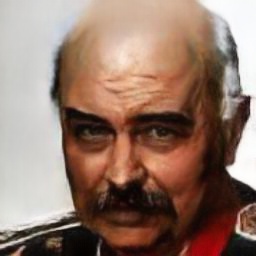} & & & &
\includegraphics[width=0.235\linewidth]{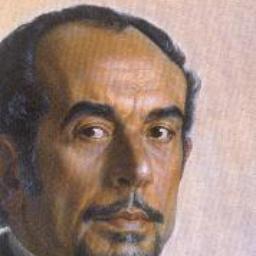} & & &
\includegraphics[width=0.235\linewidth]{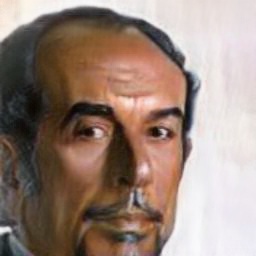} \\
\includegraphics[width=0.235\linewidth]{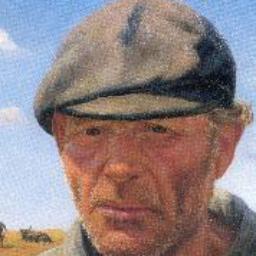}& & &
\includegraphics[width=0.235\linewidth]{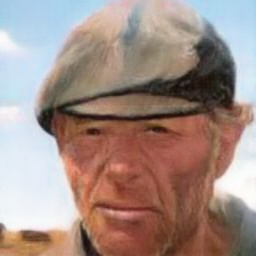} & & & &
\includegraphics[width=0.235\linewidth]{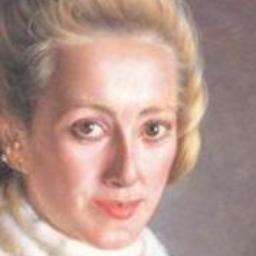} & & &
\includegraphics[width=0.235\linewidth]{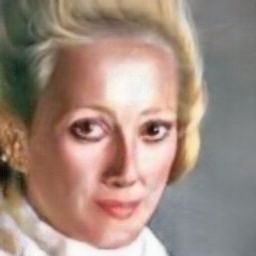} \\
\includegraphics[width=0.235\linewidth]{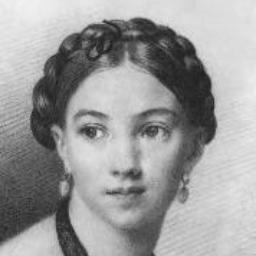}& & &
\includegraphics[width=0.235\linewidth]{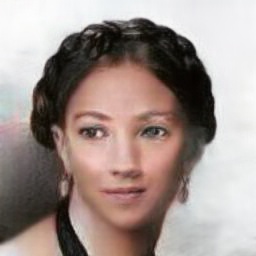} & & & &
\includegraphics[width=0.235\linewidth]{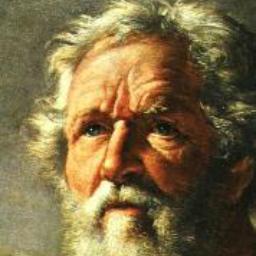} & & &
\includegraphics[width=0.235\linewidth]{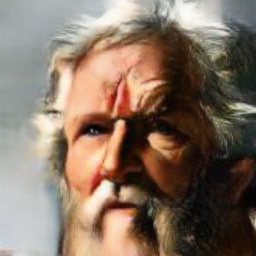} \\
\includegraphics[width=0.235\linewidth]{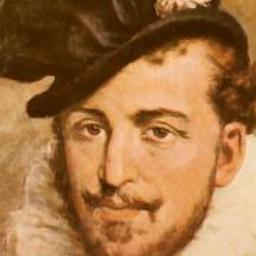}& & &
\includegraphics[width=0.235\linewidth]{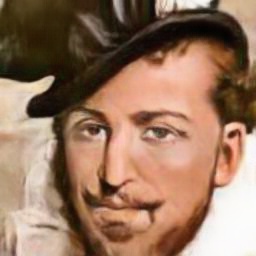} & & & &
\includegraphics[width=0.235\linewidth]{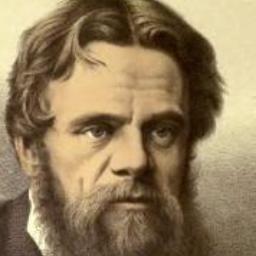} & & &
\includegraphics[width=0.235\linewidth]{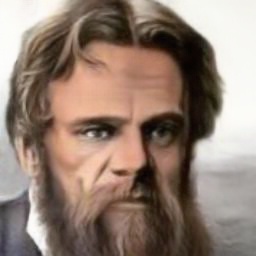} \\
\includegraphics[width=0.235\linewidth]{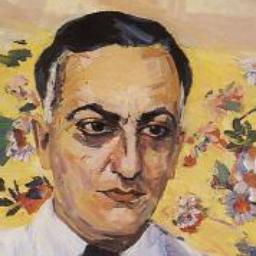}& & &
\includegraphics[width=0.235\linewidth]{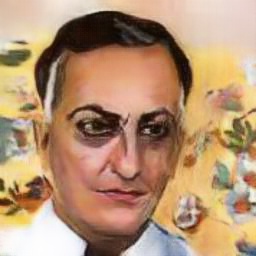} & & & &
\includegraphics[width=0.235\linewidth]{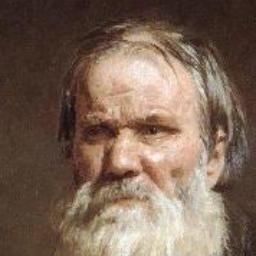} & & &
\includegraphics[width=0.235\linewidth]{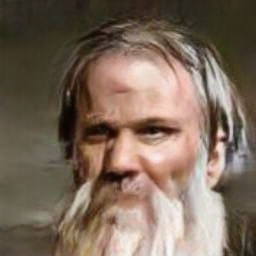} \\
\end{tabular}
\label{fig:our_results_supp2}
\end{figure*}

\begin{figure*}[tb]
\centering
\small
\setlength{\tabcolsep}{.18em}
\begin{tabular}{cccc:ccccc}
& & Original Painting & & & \textbf{\ours} & Cycle-GAN~\cite{zhu2017unpaired} & UNIT~\cite{liu2017unsupervised} & DRIT~\cite{DRIT}  \\
\rotatebox{90}{\parbox[t]{1.3in}{\hspace*{\fill}Monet\hspace*{\fill}}} & & 
\includegraphics[width=0.185\linewidth]{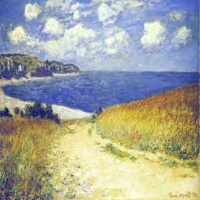}& & &
\includegraphics[width=0.185\linewidth]{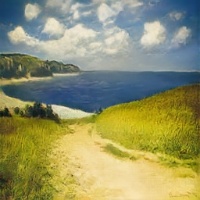} &
\includegraphics[width=0.185\linewidth]{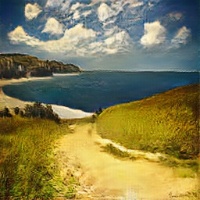} &
\includegraphics[width=0.185\linewidth]{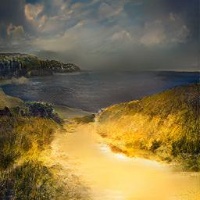} &
\includegraphics[width=0.185\linewidth]{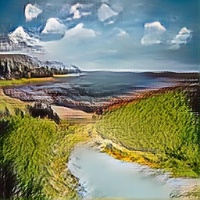} \\
\rotatebox{90}{\parbox[t]{1.3in}{\hspace*{\fill}Monet\hspace*{\fill}}} & & 
\includegraphics[width=0.185\linewidth]{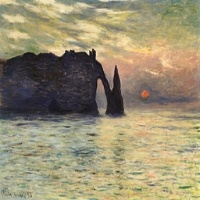}& & &
\includegraphics[width=0.185\linewidth]{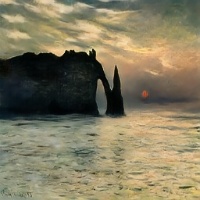} &
\includegraphics[width=0.185\linewidth]{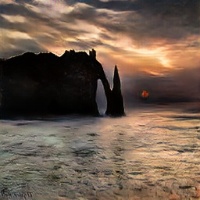} &
\includegraphics[width=0.185\linewidth]{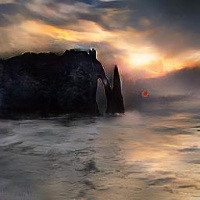} &
\includegraphics[width=0.185\linewidth]{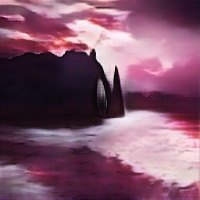} \\
\rotatebox{90}{\parbox[t]{1.3in}{\hspace*{\fill}Monet\hspace*{\fill}}} & & 
\includegraphics[width=0.185\linewidth]{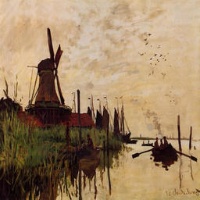}& & &
\includegraphics[width=0.185\linewidth]{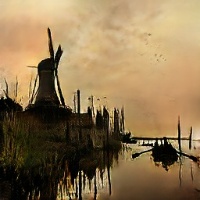} &
\includegraphics[width=0.185\linewidth]{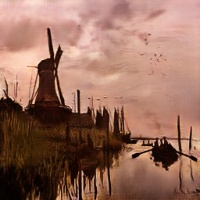} &
\includegraphics[width=0.185\linewidth]{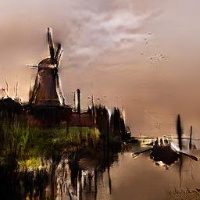} &
\includegraphics[width=0.185\linewidth]{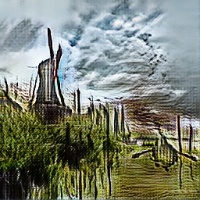} \\
\rotatebox{90}{\parbox[t]{1.3in}{\hspace*{\fill}Monet\hspace*{\fill}}} & & 
\includegraphics[width=0.185\linewidth]{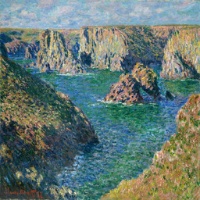}& & &
\includegraphics[width=0.185\linewidth]{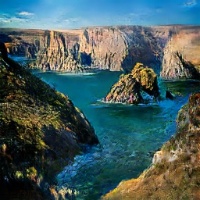} &
\includegraphics[width=0.185\linewidth]{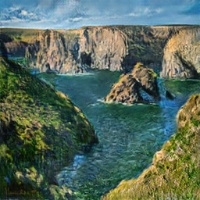} &
\includegraphics[width=0.185\linewidth]{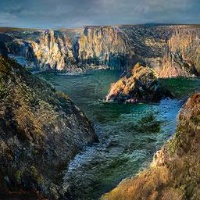} &
\includegraphics[width=0.185\linewidth]{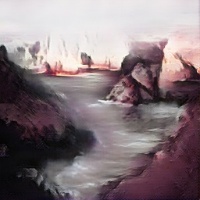} \\
\rotatebox{90}{\parbox[t]{1.3in}{\hspace*{\fill}Monet\hspace*{\fill}}} & & 
\includegraphics[width=0.185\linewidth]{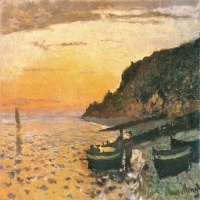}& & &
\includegraphics[width=0.185\linewidth]{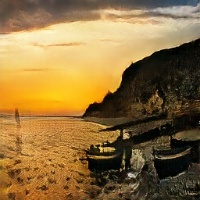} &
\includegraphics[width=0.185\linewidth]{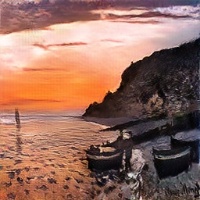} &
\includegraphics[width=0.185\linewidth]{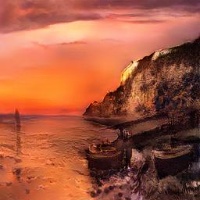} &
\includegraphics[width=0.185\linewidth]{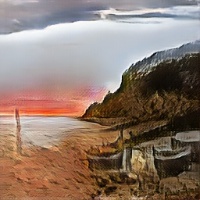} \\
\rotatebox{90}{\parbox[t]{1.3in}{\hspace*{\fill}Cezanne\hspace*{\fill}}} & & 
\includegraphics[width=0.185\linewidth]{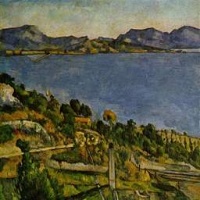}& & &
\includegraphics[width=0.185\linewidth]{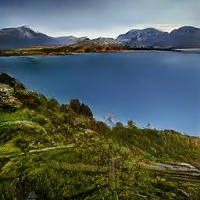} &
\includegraphics[width=0.185\linewidth]{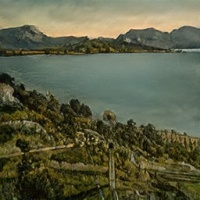} &
\includegraphics[width=0.185\linewidth]{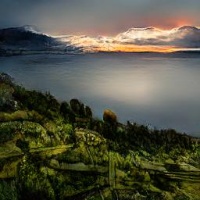} &
\includegraphics[width=0.185\linewidth]{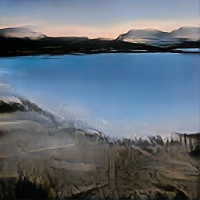} \\
\end{tabular}
\label{fig:results_supp1}
\end{figure*}

\begin{figure*}[tb]
\centering
\small
\setlength{\tabcolsep}{.18em}
\begin{tabular}{cccc:ccccc}
& & Original Painting & & & \textbf{\ours} & Cycle-GAN~\cite{zhu2017unpaired} & UNIT~\cite{liu2017unsupervised} & DRIT~\cite{DRIT}  \\
\rotatebox{90}{\parbox[t]{1.3in}{\hspace*{\fill}Cezanne\hspace*{\fill}}} & & 
\includegraphics[width=0.185\linewidth]{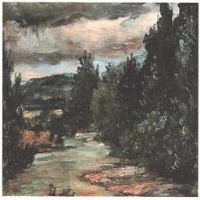}& & &
\includegraphics[width=0.185\linewidth]{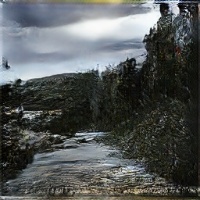} &
\includegraphics[width=0.185\linewidth]{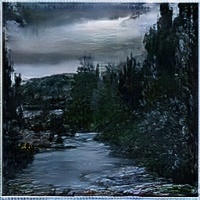} &
\includegraphics[width=0.185\linewidth]{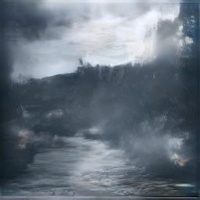} &
\includegraphics[width=0.185\linewidth]{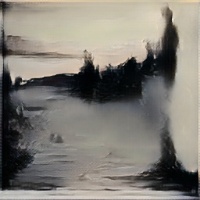} \\
\rotatebox{90}{\parbox[t]{1.3in}{\hspace*{\fill}Cezanne\hspace*{\fill}}} & & 
\includegraphics[width=0.185\linewidth]{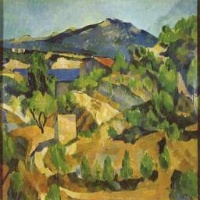}& & &
\includegraphics[width=0.185\linewidth]{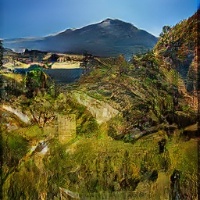} &
\includegraphics[width=0.185\linewidth]{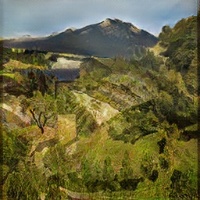} &
\includegraphics[width=0.185\linewidth]{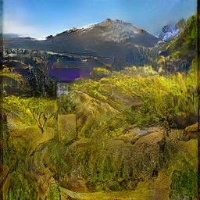} &
\includegraphics[width=0.185\linewidth]{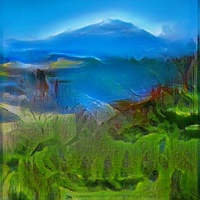} \\
\rotatebox{90}{\parbox[t]{1.3in}{\hspace*{\fill}Van Gogh\hspace*{\fill}}} & & 
\includegraphics[width=0.185\linewidth]{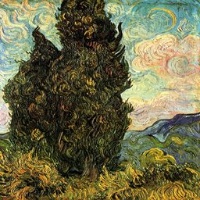}& & &
\includegraphics[width=0.185\linewidth]{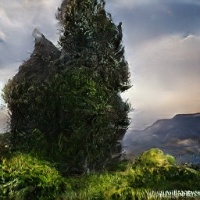} &
\includegraphics[width=0.185\linewidth]{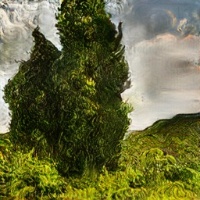} &
\includegraphics[width=0.185\linewidth]{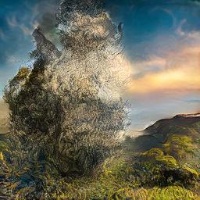} &
\includegraphics[width=0.185\linewidth]{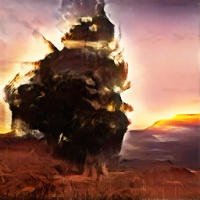} \\
\rotatebox{90}{\parbox[t]{1.3in}{\hspace*{\fill}Van Gogh\hspace*{\fill}}} & & 
\includegraphics[width=0.185\linewidth]{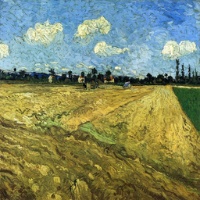}& & &
\includegraphics[width=0.185\linewidth]{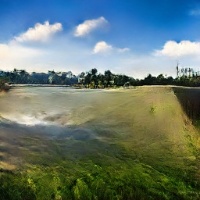} &
\includegraphics[width=0.185\linewidth]{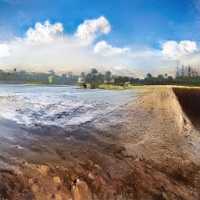} &
\includegraphics[width=0.185\linewidth]{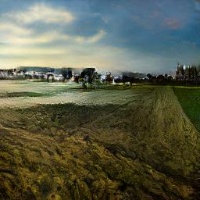} &
\includegraphics[width=0.185\linewidth]{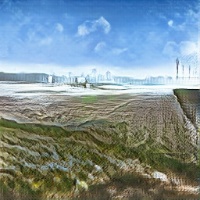} \\
\rotatebox{90}{\parbox[t]{1.3in}{\hspace*{\fill}Van Gogh\hspace*{\fill}}} & & 
\includegraphics[width=0.185\linewidth]{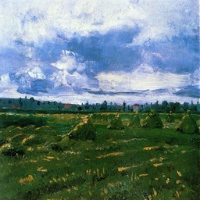}& & &
\includegraphics[width=0.185\linewidth]{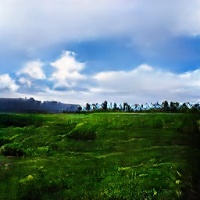} &
\includegraphics[width=0.185\linewidth]{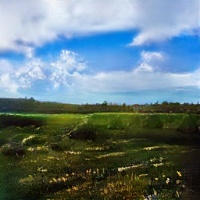} &
\includegraphics[width=0.185\linewidth]{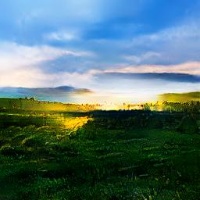} &
\includegraphics[width=0.185\linewidth]{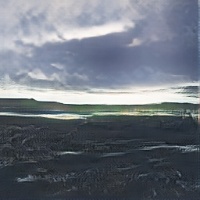} \\
\rotatebox{90}{\parbox[t]{1.3in}{\hspace*{\fill}Van Gogh\hspace*{\fill}}} & & 
\includegraphics[width=0.185\linewidth]{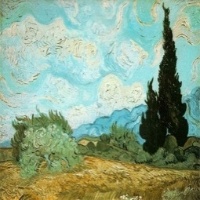}& & &
\includegraphics[width=0.185\linewidth]{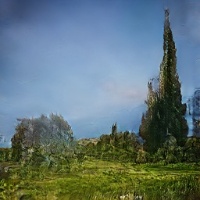} &
\includegraphics[width=0.185\linewidth]{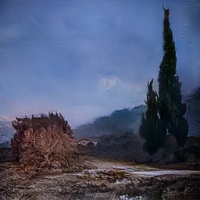} &
\includegraphics[width=0.185\linewidth]{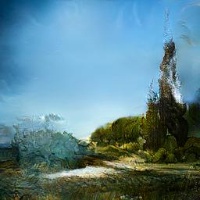} &
\includegraphics[width=0.185\linewidth]{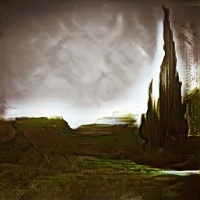} \\
\end{tabular}
\label{fig:results_supp2}
\end{figure*}

\begin{figure*}[tb]
\centering
\small
\setlength{\tabcolsep}{.18em}
\begin{tabular}{cccc:ccccc}
& & Original Painting & & & \textbf{\ours} & Cycle-GAN~\cite{zhu2017unpaired} & UNIT~\cite{liu2017unsupervised} & DRIT~\cite{DRIT}  \\
\rotatebox{90}{\parbox[t]{1.3in}{\hspace*{\fill}Ukiyo-e\hspace*{\fill}}} & & 
\includegraphics[width=0.185\linewidth]{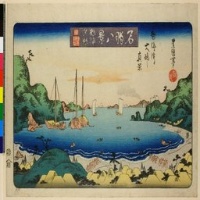}& & &
\includegraphics[width=0.185\linewidth]{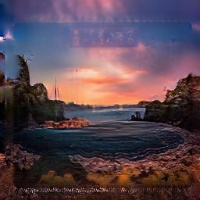} &
\includegraphics[width=0.185\linewidth]{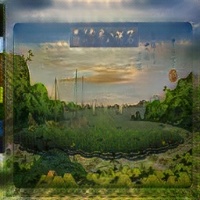} &
\includegraphics[width=0.185\linewidth]{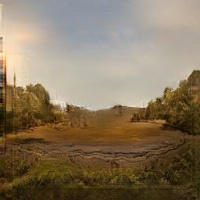} &
\includegraphics[width=0.185\linewidth]{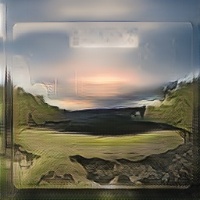} \\
\rotatebox{90}{\parbox[t]{1.3in}{\hspace*{\fill}Ukiyo-e\hspace*{\fill}}} & & 
\includegraphics[width=0.185\linewidth]{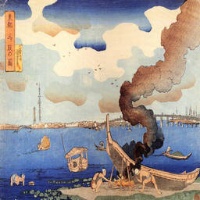}& & &
\includegraphics[width=0.185\linewidth]{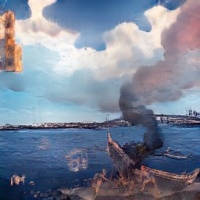} &
\includegraphics[width=0.185\linewidth]{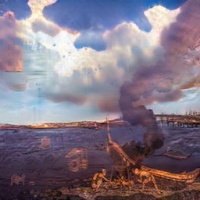} &
\includegraphics[width=0.185\linewidth]{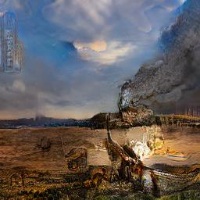} &
\includegraphics[width=0.185\linewidth]{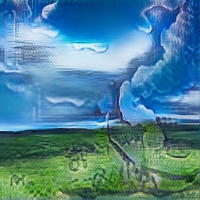} \\
\rotatebox{90}{\parbox[t]{1.3in}{\hspace*{\fill}Landscapes\hspace*{\fill}}} & & 
\includegraphics[width=0.185\linewidth]{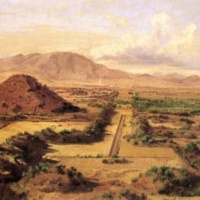}& & &
\includegraphics[width=0.185\linewidth]{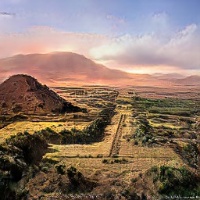} &
\includegraphics[width=0.185\linewidth]{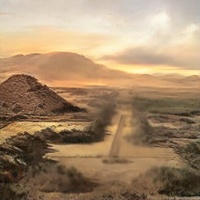} &
\includegraphics[width=0.185\linewidth]{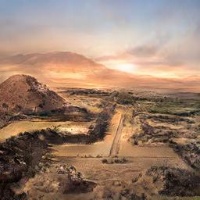} &
\includegraphics[width=0.185\linewidth]{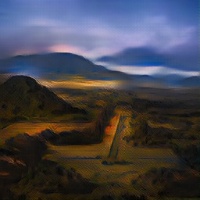} \\
\rotatebox{90}{\parbox[t]{1.3in}{\hspace*{\fill}Landscapes\hspace*{\fill}}} & & 
\includegraphics[width=0.185\linewidth]{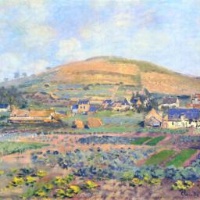}& & &
\includegraphics[width=0.185\linewidth]{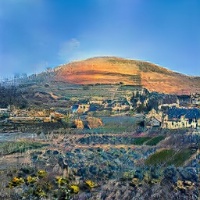} &
\includegraphics[width=0.185\linewidth]{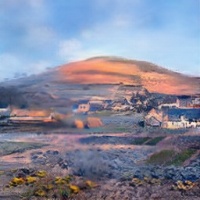} &
\includegraphics[width=0.185\linewidth]{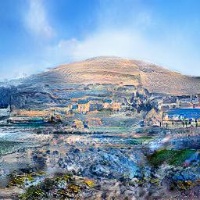} &
\includegraphics[width=0.185\linewidth]{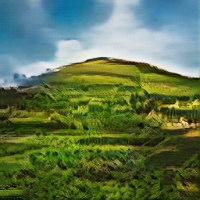} \\
\rotatebox{90}{\parbox[t]{1.3in}{\hspace*{\fill}Landscapes\hspace*{\fill}}} & & 
\includegraphics[width=0.185\linewidth]{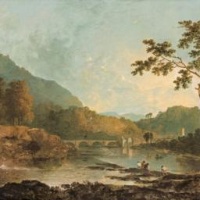}& & &
\includegraphics[width=0.185\linewidth]{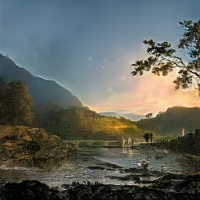} &
\includegraphics[width=0.185\linewidth]{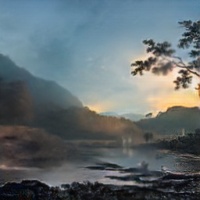} &
\includegraphics[width=0.185\linewidth]{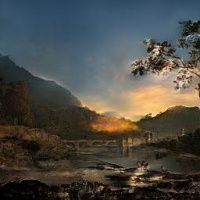} &
\includegraphics[width=0.185\linewidth]{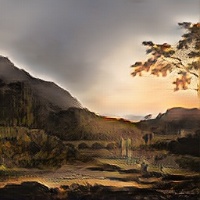} \\
\rotatebox{90}{\parbox[t]{1.3in}{\hspace*{\fill}Landscapes\hspace*{\fill}}} & & 
\includegraphics[width=0.185\linewidth]{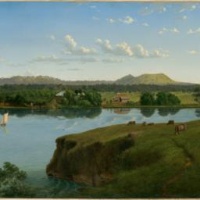}& & &
\includegraphics[width=0.185\linewidth]{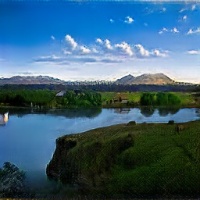} &
\includegraphics[width=0.185\linewidth]{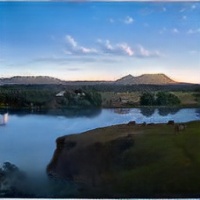} &
\includegraphics[width=0.185\linewidth]{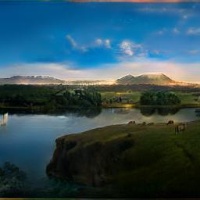} &
\includegraphics[width=0.185\linewidth]{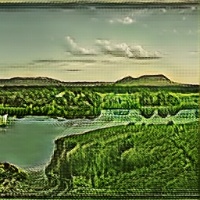} \\
\end{tabular}
\label{fig:results_supp4}
\end{figure*}

\begin{figure*}[tb]
\centering
\small
\setlength{\tabcolsep}{.18em}
\begin{tabular}{cccc:ccccc}
& & Original Painting & & & \textbf{\ours} & Cycle-GAN~\cite{zhu2017unpaired} & UNIT~\cite{liu2017unsupervised} & DRIT~\cite{DRIT}  \\
\rotatebox{90}{\parbox[t]{1.3in}{\hspace*{\fill}Landscapes\hspace*{\fill}}} & & 
\includegraphics[width=0.185\linewidth]{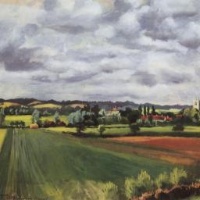}& & &
\includegraphics[width=0.185\linewidth]{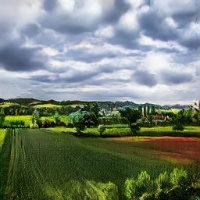} &
\includegraphics[width=0.185\linewidth]{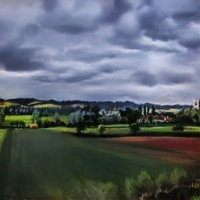} &
\includegraphics[width=0.185\linewidth]{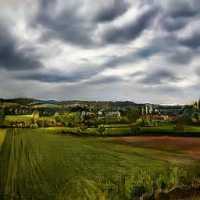} &
\includegraphics[width=0.185\linewidth]{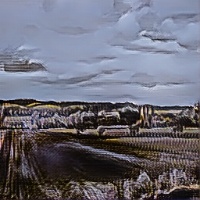} \\
\rotatebox{90}{\parbox[t]{1.3in}{\hspace*{\fill}Landscapes\hspace*{\fill}}} & & 
\includegraphics[width=0.185\linewidth]{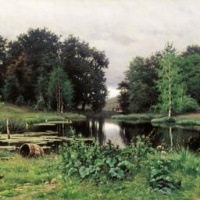}& & &
\includegraphics[width=0.185\linewidth]{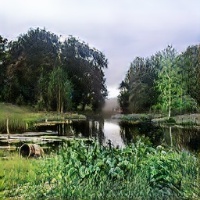} &
\includegraphics[width=0.185\linewidth]{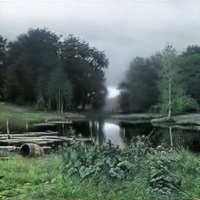} &
\includegraphics[width=0.185\linewidth]{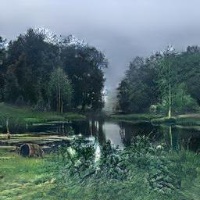} &
\includegraphics[width=0.185\linewidth]{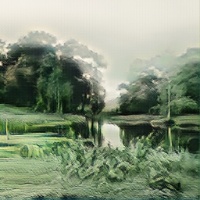} \\
\rotatebox{90}{\parbox[t]{1.3in}{\hspace*{\fill}Landscapes\hspace*{\fill}}} & & 
\includegraphics[width=0.185\linewidth]{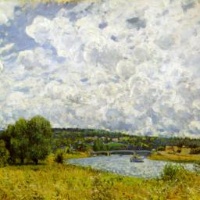}& & &
\includegraphics[width=0.185\linewidth]{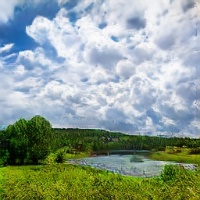} &
\includegraphics[width=0.185\linewidth]{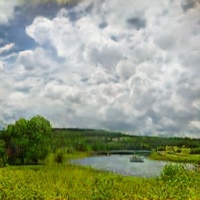} &
\includegraphics[width=0.185\linewidth]{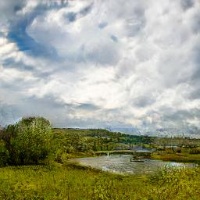} &
\includegraphics[width=0.185\linewidth]{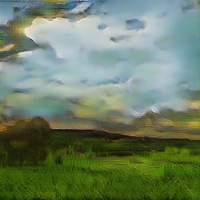} \\
\rotatebox{90}{\parbox[t]{1.3in}{\hspace*{\fill}Landscapes\hspace*{\fill}}} & & 
\includegraphics[width=0.185\linewidth]{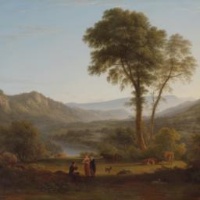}& & &
\includegraphics[width=0.185\linewidth]{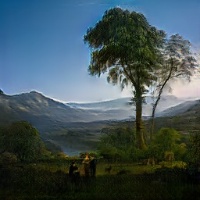} &
\includegraphics[width=0.185\linewidth]{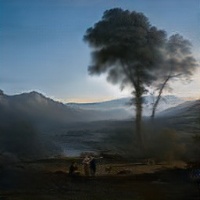} &
\includegraphics[width=0.185\linewidth]{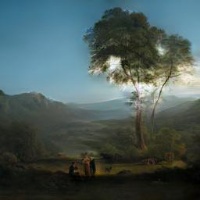} &
\includegraphics[width=0.185\linewidth]{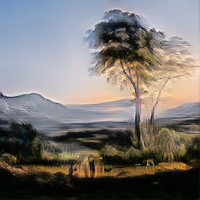} \\
\rotatebox{90}{\parbox[t]{1.3in}{\hspace*{\fill}Landscapes\hspace*{\fill}}} & & 
\includegraphics[width=0.185\linewidth]{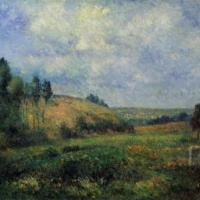}& & &
\includegraphics[width=0.185\linewidth]{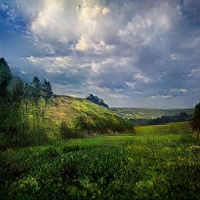} &
\includegraphics[width=0.185\linewidth]{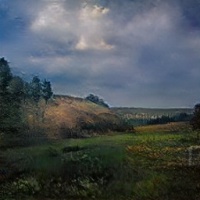} &
\includegraphics[width=0.185\linewidth]{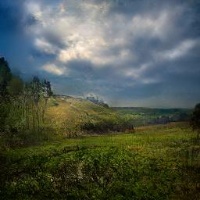} &
\includegraphics[width=0.185\linewidth]{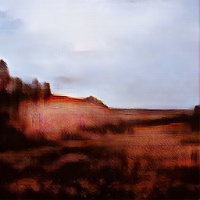} \\
\rotatebox{90}{\parbox[t]{1.3in}{\hspace*{\fill}Landscapes\hspace*{\fill}}} & & 
\includegraphics[width=0.185\linewidth]{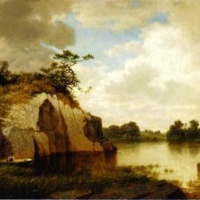}& & &
\includegraphics[width=0.185\linewidth]{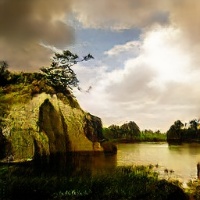} &
\includegraphics[width=0.185\linewidth]{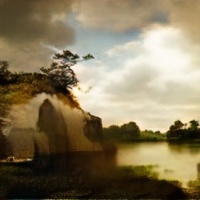} &
\includegraphics[width=0.185\linewidth]{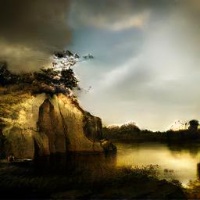} &
\includegraphics[width=0.185\linewidth]{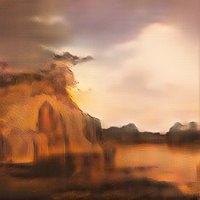} \\
\end{tabular}
\label{fig:results_supp5}
\end{figure*}

\begin{figure*}[tb]
\centering
\small
\setlength{\tabcolsep}{.18em}
\begin{tabular}{cccc:ccccc}
& & Original Painting & & & \textbf{\ours} & Cycle-GAN~\cite{zhu2017unpaired} & UNIT~\cite{liu2017unsupervised} & DRIT~\cite{DRIT}  \\
\rotatebox{90}{\parbox[t]{1.3in}{\hspace*{\fill}Landscapes\hspace*{\fill}}} & & 
\includegraphics[width=0.185\linewidth]{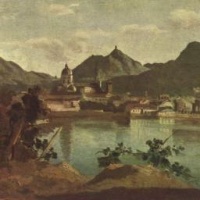}& & &
\includegraphics[width=0.185\linewidth]{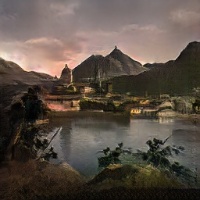} &
\includegraphics[width=0.185\linewidth]{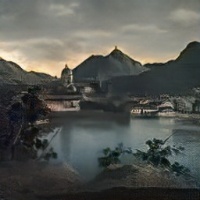} &
\includegraphics[width=0.185\linewidth]{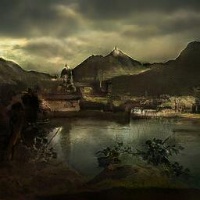} &
\includegraphics[width=0.185\linewidth]{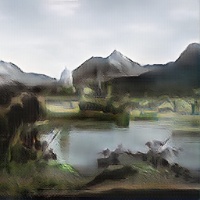} \\
\rotatebox{90}{\parbox[t]{1.3in}{\hspace*{\fill}Landscapes\hspace*{\fill}}} & & 
\includegraphics[width=0.185\linewidth]{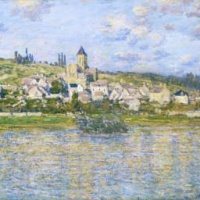}& & &
\includegraphics[width=0.185\linewidth]{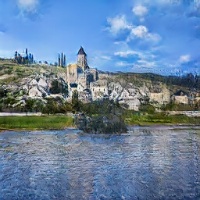} &
\includegraphics[width=0.185\linewidth]{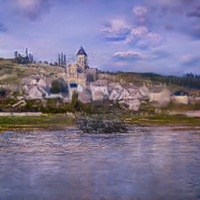} &
\includegraphics[width=0.185\linewidth]{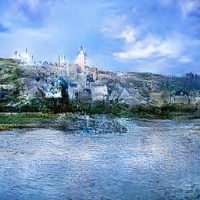} &
\includegraphics[width=0.185\linewidth]{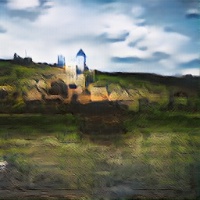} \\
\rotatebox{90}{\parbox[t]{1.3in}{\hspace*{\fill}Landscapes\hspace*{\fill}}} & & 
\includegraphics[width=0.185\linewidth]{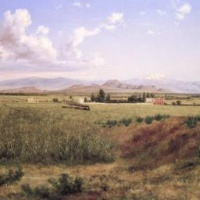}& & &
\includegraphics[width=0.185\linewidth]{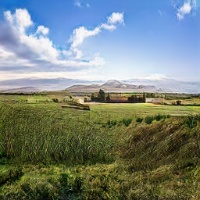} &
\includegraphics[width=0.185\linewidth]{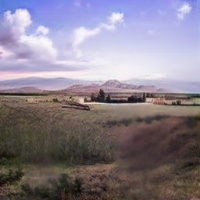} &
\includegraphics[width=0.185\linewidth]{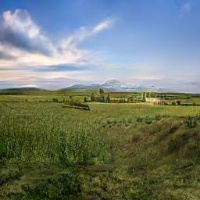} &
\includegraphics[width=0.185\linewidth]{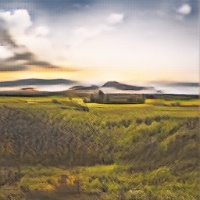} \\
\rotatebox{90}{\parbox[t]{1.3in}{\hspace*{\fill}Landscapes\hspace*{\fill}}} & & 
\includegraphics[width=0.185\linewidth]{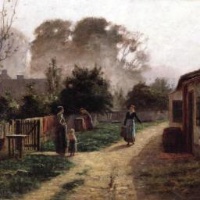}& & &
\includegraphics[width=0.185\linewidth]{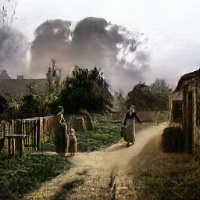} &
\includegraphics[width=0.185\linewidth]{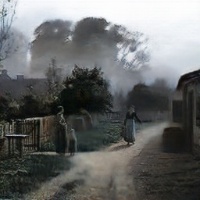} &
\includegraphics[width=0.185\linewidth]{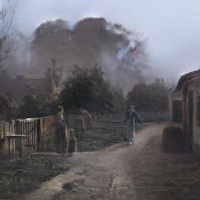} &
\includegraphics[width=0.185\linewidth]{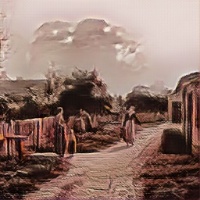} \\
\rotatebox{90}{\parbox[t]{1.3in}{\hspace*{\fill}Landscapes\hspace*{\fill}}} & & 
\includegraphics[width=0.185\linewidth]{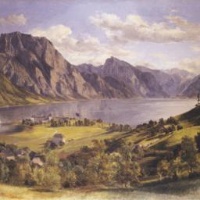}& & &
\includegraphics[width=0.185\linewidth]{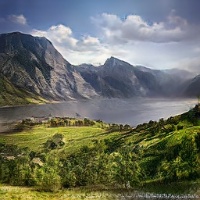} &
\includegraphics[width=0.185\linewidth]{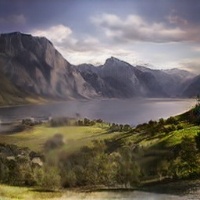} &
\includegraphics[width=0.185\linewidth]{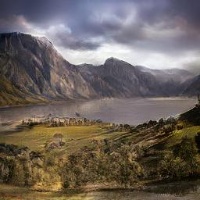} &
\includegraphics[width=0.185\linewidth]{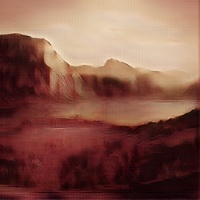} \\
\rotatebox{90}{\parbox[t]{1.3in}{\hspace*{\fill}Landscapes\hspace*{\fill}}} & & 
\includegraphics[width=0.185\linewidth]{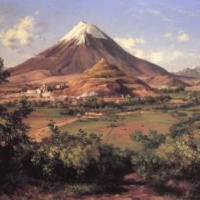}& & &
\includegraphics[width=0.185\linewidth]{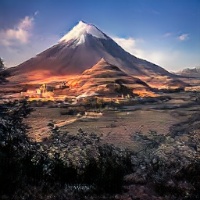} &
\includegraphics[width=0.185\linewidth]{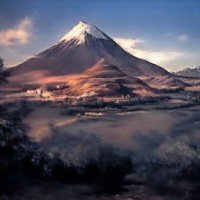} &
\includegraphics[width=0.185\linewidth]{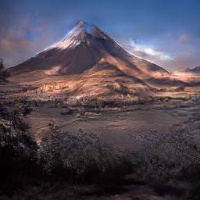} &
\includegraphics[width=0.185\linewidth]{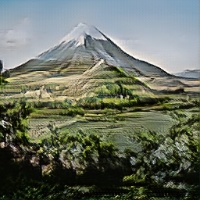} \\
\end{tabular}
\label{fig:results_supp6}
\end{figure*}

\begin{figure*}[tb]
\centering
\small
\setlength{\tabcolsep}{.18em}
\begin{tabular}{cccc:ccccc}
& & Original Painting & & & \textbf{\ours} & Cycle-GAN~\cite{zhu2017unpaired} & UNIT~\cite{liu2017unsupervised} & DRIT~\cite{DRIT}  \\
\rotatebox{90}{\parbox[t]{1.3in}{\hspace*{\fill}Portraits\hspace*{\fill}}} & & 
\includegraphics[width=0.185\linewidth]{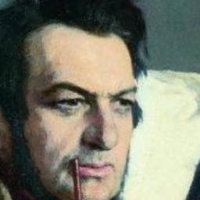}& & &
\includegraphics[width=0.185\linewidth]{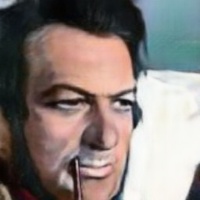} &
\includegraphics[width=0.185\linewidth]{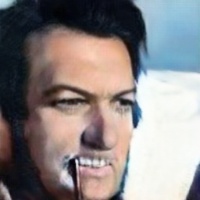} &
\includegraphics[width=0.185\linewidth]{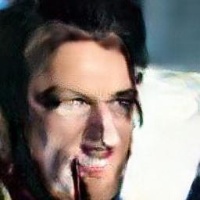} &
\includegraphics[width=0.185\linewidth]{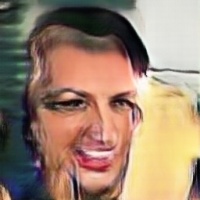} \\
\rotatebox{90}{\parbox[t]{1.3in}{\hspace*{\fill}Portraits\hspace*{\fill}}} & & 
\includegraphics[width=0.185\linewidth]{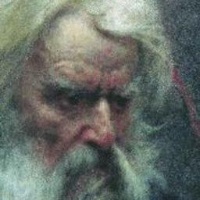}& & &
\includegraphics[width=0.185\linewidth]{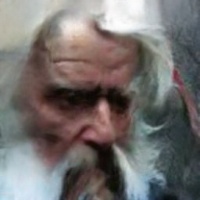} &
\includegraphics[width=0.185\linewidth]{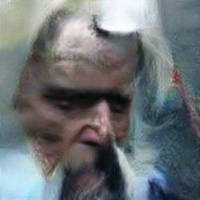} &
\includegraphics[width=0.185\linewidth]{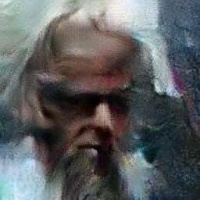} &
\includegraphics[width=0.185\linewidth]{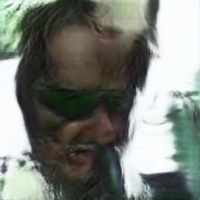} \\
\rotatebox{90}{\parbox[t]{1.3in}{\hspace*{\fill}Portraits\hspace*{\fill}}} & & 
\includegraphics[width=0.185\linewidth]{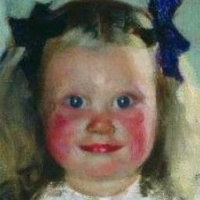}& & &
\includegraphics[width=0.185\linewidth]{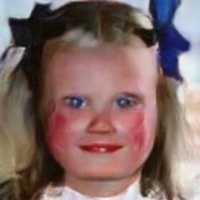} &
\includegraphics[width=0.185\linewidth]{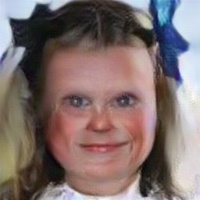} &
\includegraphics[width=0.185\linewidth]{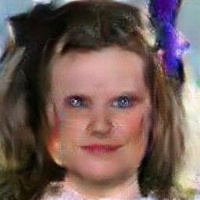} &
\includegraphics[width=0.185\linewidth]{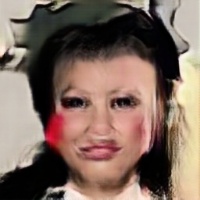} \\
\rotatebox{90}{\parbox[t]{1.3in}{\hspace*{\fill}Portraits\hspace*{\fill}}} & & 
\includegraphics[width=0.185\linewidth]{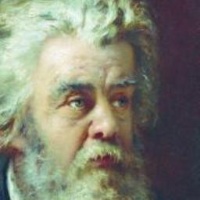}& & &
\includegraphics[width=0.185\linewidth]{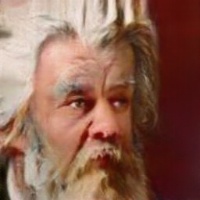} &
\includegraphics[width=0.185\linewidth]{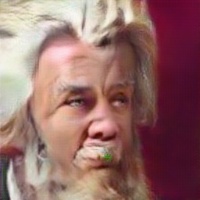} &
\includegraphics[width=0.185\linewidth]{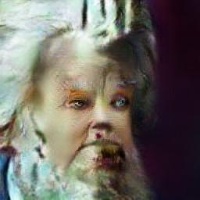} &
\includegraphics[width=0.185\linewidth]{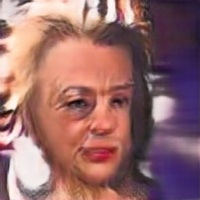} \\
\rotatebox{90}{\parbox[t]{1.3in}{\hspace*{\fill}Portraits\hspace*{\fill}}} & & 
\includegraphics[width=0.185\linewidth]{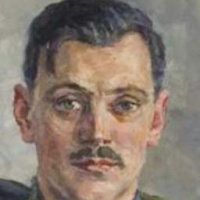}& & &
\includegraphics[width=0.185\linewidth]{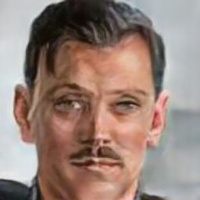} &
\includegraphics[width=0.185\linewidth]{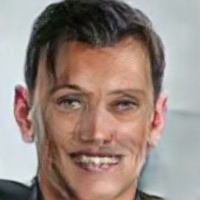} &
\includegraphics[width=0.185\linewidth]{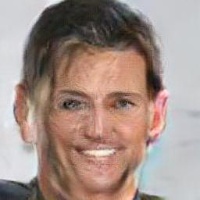} &
\includegraphics[width=0.185\linewidth]{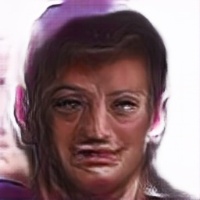} \\
\rotatebox{90}{\parbox[t]{1.3in}{\hspace*{\fill}Portraits\hspace*{\fill}}} & & 
\includegraphics[width=0.185\linewidth]{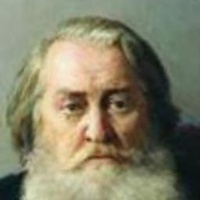}& & &
\includegraphics[width=0.185\linewidth]{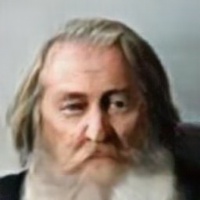} &
\includegraphics[width=0.185\linewidth]{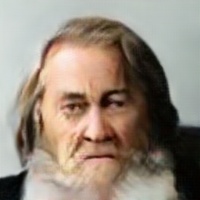} &
\includegraphics[width=0.185\linewidth]{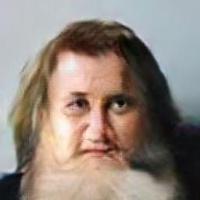} &
\includegraphics[width=0.185\linewidth]{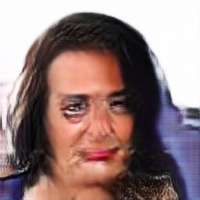} \\
\end{tabular}
\label{fig:results_supp7}
\end{figure*}

\begin{figure*}[tb]
\centering
\small
\setlength{\tabcolsep}{.18em}
\begin{tabular}{cccc:ccccc}
& & Original Painting & & & \textbf{\ours} & Cycle-GAN~\cite{zhu2017unpaired} & UNIT~\cite{liu2017unsupervised} & DRIT~\cite{DRIT}  \\

\rotatebox{90}{\parbox[t]{1.3in}{\hspace*{\fill}Portraits\hspace*{\fill}}} & & 
\includegraphics[width=0.185\linewidth]{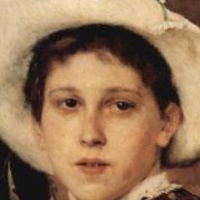}& & &
\includegraphics[width=0.185\linewidth]{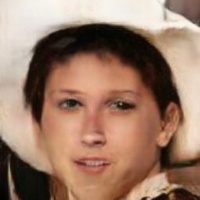} &
\includegraphics[width=0.185\linewidth]{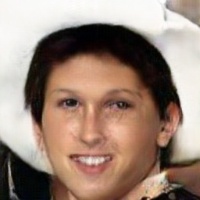} &
\includegraphics[width=0.185\linewidth]{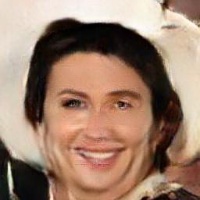} &
\includegraphics[width=0.185\linewidth]{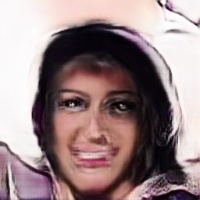} \\
\rotatebox{90}{\parbox[t]{1.3in}{\hspace*{\fill}Portraits\hspace*{\fill}}} & & 
\includegraphics[width=0.185\linewidth]{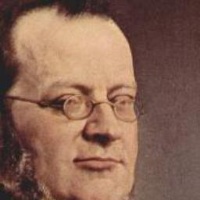}& & &
\includegraphics[width=0.185\linewidth]{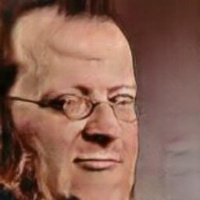} &
\includegraphics[width=0.185\linewidth]{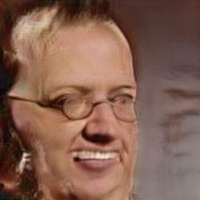} &
\includegraphics[width=0.185\linewidth]{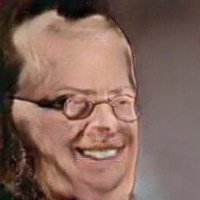} &
\includegraphics[width=0.185\linewidth]{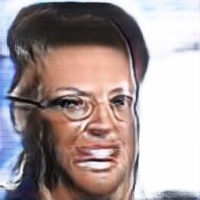} \\
\rotatebox{90}{\parbox[t]{1.3in}{\hspace*{\fill}Portraits\hspace*{\fill}}} & & 
\includegraphics[width=0.185\linewidth]{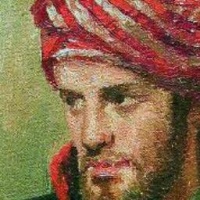}& & &
\includegraphics[width=0.185\linewidth]{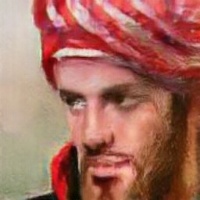} &
\includegraphics[width=0.185\linewidth]{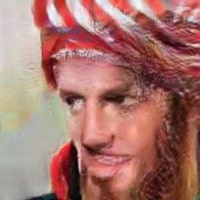} &
\includegraphics[width=0.185\linewidth]{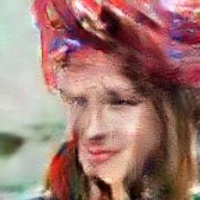} &
\includegraphics[width=0.185\linewidth]{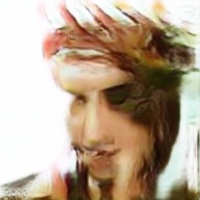} \\
\rotatebox{90}{\parbox[t]{1.3in}{\hspace*{\fill}Portraits\hspace*{\fill}}} & & 
\includegraphics[width=0.185\linewidth]{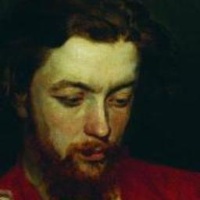}& & &
\includegraphics[width=0.185\linewidth]{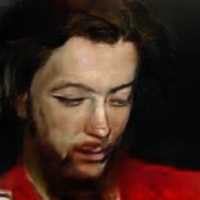} &
\includegraphics[width=0.185\linewidth]{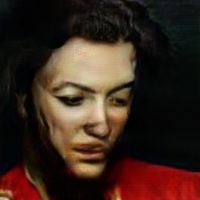} &
\includegraphics[width=0.185\linewidth]{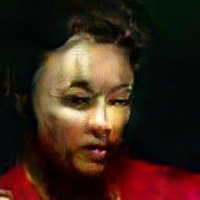} &
\includegraphics[width=0.185\linewidth]{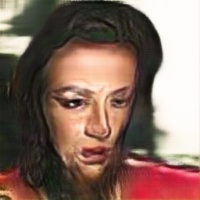} \\
\rotatebox{90}{\parbox[t]{1.3in}{\hspace*{\fill}Portraits\hspace*{\fill}}} & & 
\includegraphics[width=0.185\linewidth]{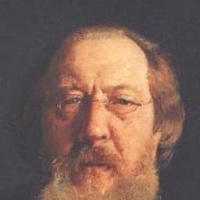}& & &
\includegraphics[width=0.185\linewidth]{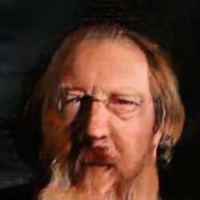} &
\includegraphics[width=0.185\linewidth]{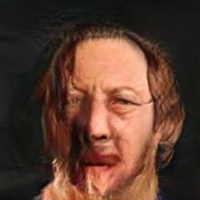} &
\includegraphics[width=0.185\linewidth]{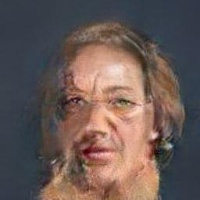} &
\includegraphics[width=0.185\linewidth]{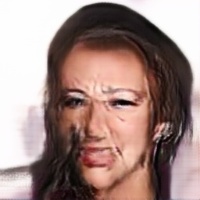} \\
\rotatebox{90}{\parbox[t]{1.3in}{\hspace*{\fill}Portraits\hspace*{\fill}}} & & 
\includegraphics[width=0.185\linewidth]{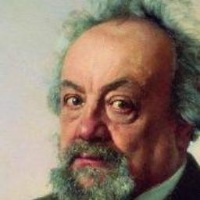}& & &
\includegraphics[width=0.185\linewidth]{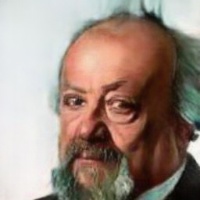} &
\includegraphics[width=0.185\linewidth]{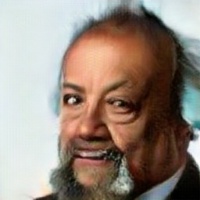} &
\includegraphics[width=0.185\linewidth]{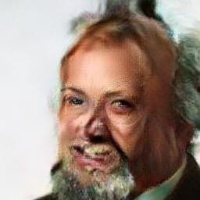} &
\includegraphics[width=0.185\linewidth]{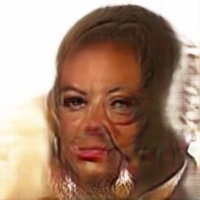} \\
\end{tabular}
\label{fig:results_supp8}
\end{figure*}

\end{document}